\RequirePackage{fix-cm}
\documentclass[twocolumn]{svjour3}          
\smartqed  
\usepackage{graphicx}

\usepackage{times}
\usepackage[ruled]{algorithm2e} 
\usepackage{epsfig}
\usepackage{amsmath}
\usepackage{amssymb}
\usepackage{mathrsfs}
\usepackage{url}
\usepackage{enumitem}
\usepackage{algpseudocode}
\usepackage{dsfont}
\usepackage{booktabs}
\usepackage{multirow}
\usepackage{enumitem}
\usepackage[pagebackref=false,breaklinks=true,letterpaper=true,colorlinks,urlcolor=blue,citecolor=blue,linkcolor=blue,bookmarks=false]{hyperref}

\journalname{International Journal of Computer Vision (IJCV)}
\begin{document}

\title{Unsupervised Deep Representation Learning for Real-Time Tracking
}
\subtitle{}


\author{Ning Wang \and Wengang Zhou \and Yibing Song \and Chao Ma \and Wei Liu \and
	Houqiang Li 
}


\institute{
	Ning Wang \at
	The CAS Key Laboratory of GIPAS, University of Science and Technology of China, Hefei, China.\\
	\email{wn6149@mail.ustc.edu.cn}  
	\and
	Wengang Zhou \at
	The CAS Key Laboratory of GIPAS, University of Science and Technology of China, Hefei, China.\\
	Institute of Artificial Intelligence, Hefei Comprehensive National Science Center, Hefei, China.\\
	\email{zhwg@ustc.edu.cn}  
	\and
	Yibing Song \at
	Tencent AI Lab, Shenzhen, China.\\
	\email{yibingsong.cv@gmail.com}  
	\and 
	Chao Ma \at
	The MoE Key Lab of Artificial Intelligence, AI Institute, Shanghai Jiao Tong University, Shanghai, China.\\
	\email{chaoma@sjtu.edu.cn}  
	\and
	Wei Liu \at
	Tencent AI Lab, Shenzhen, China.\\
	\email{wl2223@columbia.edu} 
	\and
	Houqiang Li \at
	The CAS Key Laboratory of GIPAS, University of Science and Technology of China, Hefei, China.\\
	Institute of Artificial Intelligence, Hefei Comprehensive National Science Center, Hefei, China.\\
	\email{lihq@ustc.edu.cn}  
	\and
	Corresponding Authors: Wengang Zhou and Houqiang Li
}

\date{Received: date / Accepted: date}

\maketitle

\sloppy{}

\begin{abstract}
The advancement of visual tracking has continuously been brought by deep learning models. 
Typically, supervised learning is employed to train these models with expensive labeled data.
In order to reduce the workload of manual annotations and learn to track arbitrary objects, we propose an unsupervised learning method for visual tracking.
The motivation of our unsupervised learning is that a robust tracker should be effective in bidirectional tracking. Specifically, the tracker is able to forward localize a target object in successive frames and backtrace to its initial position in the first frame.
Based on such a motivation, in the training process, we measure the consistency between forward and backward trajectories to learn a robust tracker from scratch merely using unlabeled videos.
We build our framework on a Siamese correlation filter network, and propose a multi-frame validation scheme and a cost-sensitive loss to facilitate unsupervised learning.		
Without bells and whistles, the proposed unsupervised tracker achieves the baseline accuracy as classic fully supervised trackers while achieving a real-time speed.
Furthermore, our unsupervised framework exhibits a potential in leveraging more unlabeled or weakly labeled data to further improve the tracking accuracy.

\end{abstract}

\section{Introduction}

Visual object tracking is a fundamental task in computer vision with numerous applications including video surveillance, autonomous driving, augmented reality, and human-computer interactions. It aims to localize a moving object annotated at the initial frame with a bounding box.
Recently, deep models have improved the tracking accuracies by strengthening the feature representations~\cite{HCF,C-COT,ECO} or optimizing networks end-to-end~\cite{SiamFc,SiamRPN,MDNet,CFNet}.
These models are offline pretrained with full supervision, which requires a large number of annotated ground-truth labels during the training stage.
Manual annotations are always expensive and time-consuming, whereas a huge number of unlabeled videos are readily available on the Internet.
On the other hand, visual tracking differs from other recognition tasks (e.g., object detection, image classification) in the sense that object labels vary according to target initializations on the first frame.
%
%
The extensive and uncertain labeling process for supervised learning raises our interest to develop an alternative learning scheme by using unlabeled video sequences in the wild.

\begin{figure}
	\centering
	\includegraphics[width=8.6cm]{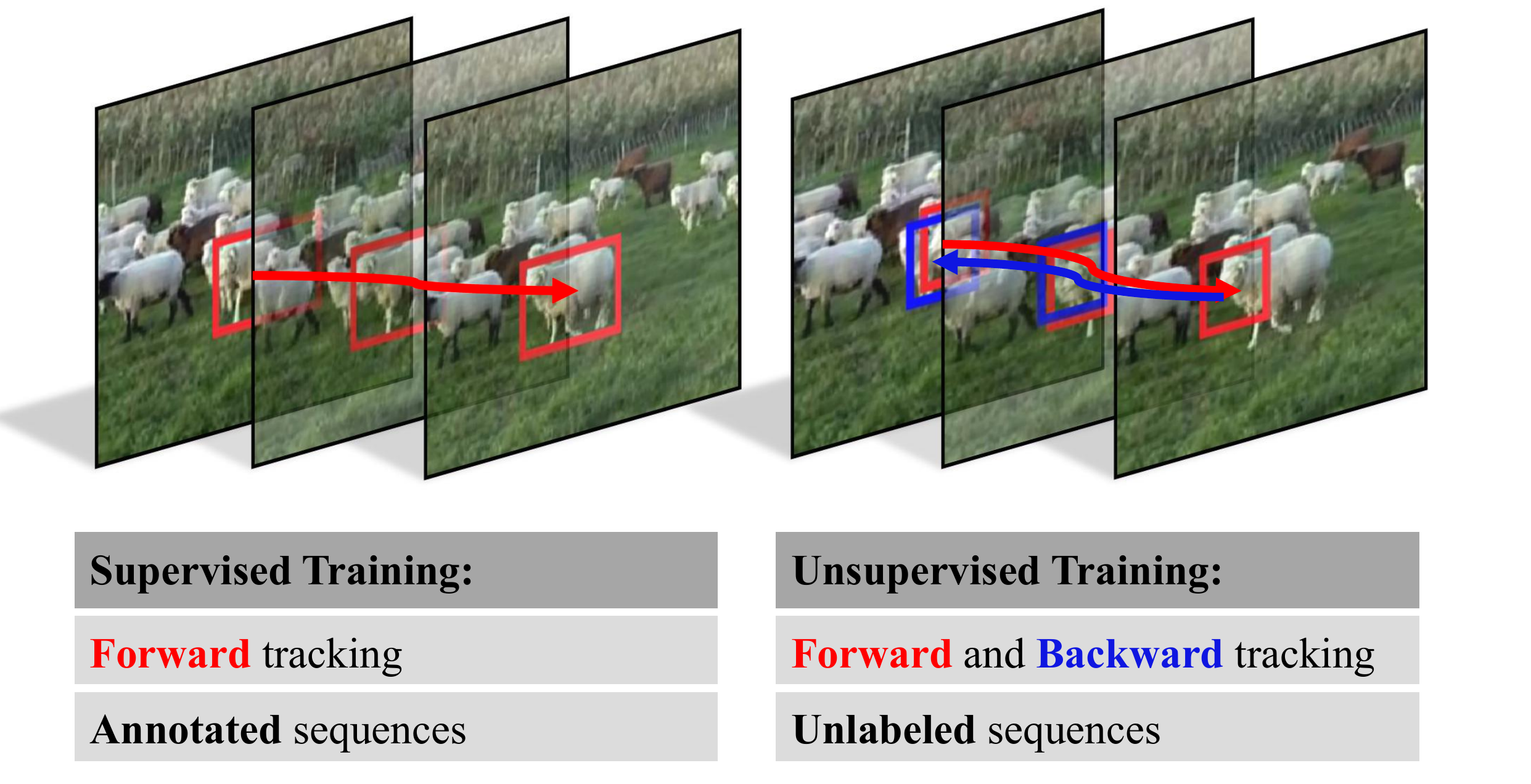}
	\caption{Visual tracking via supervised and unsupervised learnings. Supervised learning requires ground-truth labels for individual frames in the training videos, while our proposed unsupervised learning is free of any labels by measuring the trajectory consistency between forward and backward tracking.}
	\label{fig:introduction} 
\end{figure}

In this paper, we propose an unsupervised learning approach for visual tracking.
Instead of using off-the-shelf deep models, we train the visual tracking network from scratch.
The intuition of unsupervised learning resides on the bidirectional motion analysis in video sequences.
Tracking an object can be executed in both the forward and backward ways.
Initially, given the bounding box annotation of a target object in the first frame, we can track the target object forward in the subsequent frames.
When tracking backward, we use the predicted location in the last frame as the initial target bounding box, and track it backward towards the first frame.
Ideally, the estimated bounding box location in the first frame is identical with the given one in the forward pass.
In this work, we measure the difference between the forward and backward target trajectories and formulate it as a loss function.
We use the computed loss to train our network in a self-supervised manner\footnote{In this paper, we do not distinguish between the terms \emph{unsupervised} and \emph{self-supervised}, as both refer to learning without ground-truth annotations.}, as shown in Fig. \ref{fig:introduction}.
By repeatedly tracking forward and backward, our model learns to locate target objects in consecutive frames without any supervision.

The proposed unsupervised training aims to learn a generic feature representation instead of strictly focusing on tracking a complete object.
In the first frame, we initialize a bounding box that covers the informative local region with high image entropy.
The bounding box may contain arbitrary image content and may not cover an entire object.
Then, our tracking network learns to track the bounding box region in the training video sequences.
Our unsupervised annotation shares similarity with the part-based \cite{Liu_StructuralCF} and edge-based \cite{IBCCF} tracking methods that track the subregions of a target object.
We expect our tracker not only to concentrate on the shape of a complete object, but also to track any part of it.
The bounding box initialization by image entropy gets rid of the manual annotation on the first frame and thus ensures the whole learning process unsupervised.

We employ unsupervised learning under the Siamese correlation filter framework.
The training steps consist of forward tracking and backward verification.
A limitation of the forward and backward consistency measurement is that the target trajectory in the forward pass may coincide with that in the backward pass although the tracker loses the target.
The consistency loss function fails to penalize this situation because the predicted target region can still backtrace to the initial position on the first frame regardless of losing the target.
In addition, challenges such as heavy occlusion or out-of-view in training videos will degrade the CNN feature representation capability.
To tackle these issues, we introduce a multi-frame validation scheme and a cost-sensitive loss to facilitate unsupervised training.
If the tracker loses the target, the trajectories predicted from the forward and backward directions are unlikely to be consistent when more frames are used in the training stage. Besides, we propose a new cost-sensitive loss to alleviate the impact of the noisy samples during unsupervised learning.
The training samples containing background texture will be excluded by the image entropy measurement.
Based on the multi-frame validation and sample selection strategies discussed above, our network training is stabilized.

We evaluate our method on the challenging benchmark datasets including OTB-2013 \cite{OTB-2013}, OTB-2015 \cite{OTB-2015}, Temple-Color \cite{TempleColor128}, VOT2016 \cite{VOT2016}, VOT2017/2018 \cite{VOT2018}, LaSOT \cite{LaSOT}, and TrackingNet \cite{2018trackingnet}.
Extensive experimental results indicate that without bells and whistles, the proposed unsupervised tracker is even comparable with the baseline configuration of fully supervised trackers \cite{SiamFc,CFNet,DCFNet}.
When integrated with an adaptive online model update \cite{SRDCFdecon,ECO}, the proposed tracker shows state-of-the-art performance.
It is worth mentioning that our tracker trained via unsupervised learning achieves comparable performance with that via supervised learning when only limited or noisy labels are available.
%
In addition, we demonstrate the potential of our tracker to further boost the accuracy by using more unlabeled data.
%
%
A complete analysis of various training configurations is given in Section \ref{ablation}.

In summary, the contributions of this work are three-fold:
\begin{itemize}
	\setlength{\parskip}{0pt}
	\item We propose an unsupervised learning method on the Siamese correlation filter network. The unsupervised learning consists of forward and backward trackings to measure the trajectory consistency for network training.
	\item We propose a multi-frame validation scheme to enlarge the trajectory inconsistency when the tracker loses the target. In addition, we propose a cost-sensitive loss and an entropy selection metric to reduce the contributions from easy samples in the training process.
	\item The extensive experiments carried out on seven standard benchmarks show the favorable performance of the proposed tracker. We provide an in-depth analysis of our unsupervised representation and reveal the potential of unsupervised learning in visual tracking.
\end{itemize}

In the remainder of this paper, we describe the related work in Section \ref{sec:related work}, the proposed method in Section \ref{sec:proposed approach}, and the experiments in Section \ref{experiment}. Finally, we conclude the paper in Section \ref{conclusion}.

\section{Related Work}\label{sec:related work}

In this section, we perform a literature review on deep tracking methods, forward-backward motion analysis, and unsupervised representation learning.

\subsection{Deep Visual Tracking}
Deep models have influenced visual tracking mainly from two perspectives. The first one is to provide a discriminative CNN feature representation by using off-the-shelf backbones (e.g., VGG \cite{VGG,VGGM}), while the second one is to formulate a complete tracking network for end-to-end training and predictions. The discriminative correlation filters (DCFs) \cite{MOSSE,KCF,DSST,huang2017applyingIJCV,LCT_IJCV,exploitingAnisotropy,CSRDCF_IJCV} handle the visual tracking task by solving a ridge regression task using densely sampled candidates. While being integrated with discriminative CNN features, the remaining operations (e.g., regression solver, online update) are kept still in the DCF trackers \cite{C-COT,STRCF,MCCT,ECO}. On the other hand, the end-to-end learning network can be categorized as classification and regression based networks. The classification networks \cite{MDNet,VITAL,RTMDNet,DAT_Pu} incrementally train a binary classifier to differentiate the target and background distractors. The regression networks \cite{CREST,DSLT,luo2019end} use CNN layers to regress CNN features of the search region to a response map for accurate localization. These end-to-end learning networks need online update and inevitably increase the computational burden.

Recently, the Siamese network has received huge investigations because of its efficiency in online prediction. The SiamFC tracker \cite{SiamFc} uses a cross-correlation layer to measure feature similarity between the template patch and search patches. The fully convolutional nature of SiamFC efficiently predicts the target response without redundancy. By incorporating the region proposal network (RPN) \cite{ren-pami16-faster}, the SiamRPN methods \cite{SiamRPN,DaSiamRPN} achieve state-of-the-art performance while running at 160 FPS. Other improvements based on Siamese networks include ensemble learning \cite{SASiam}, dynamic memory \cite{MemTrack}, attention modulation \cite{RASNet}, capacity increments \cite{deeperwiderSiamFC}, and reinforcement learning \cite{EAST,HP}.
By integrating the correlation filter, the Siamese correlation filter network \cite{CFNet,DCFNet} achieves favorable performance even with an extremely lightweight model.
Different from the above deep trackers that train a CNN model in a supervised manner or directly use off-the-shelf deep models, we propose to learn a tracking network from scratch using unlabeled videos via unsupervised learning.

\begin{figure*}
	\centering
	\includegraphics[width=17.3cm]{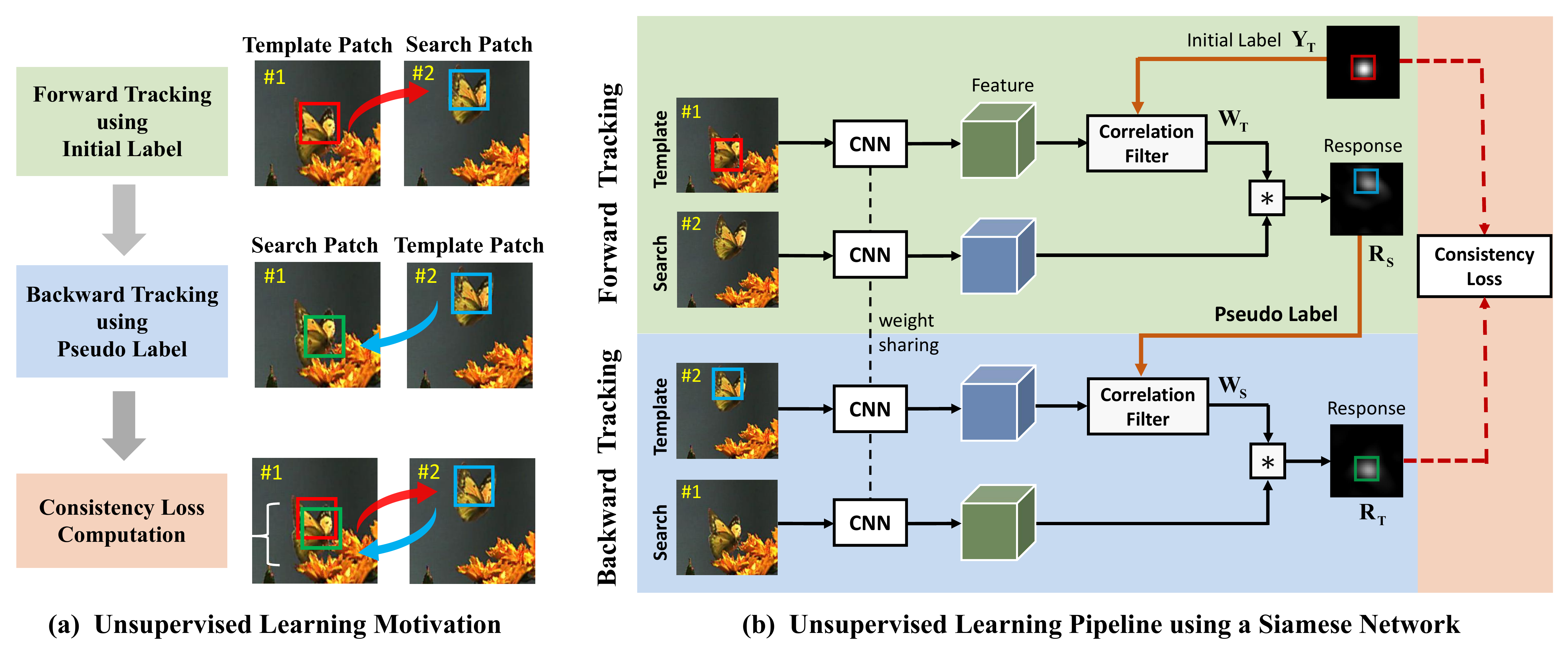}
	\caption{An overview of unsupervised learning in deep tracking. We show our motivation in (a) that we track forward and backward to compute the consistency loss for network training. The detailed training procedure is shown in (b), where unsupervised learning is integrated into a Siamese correlation filter network. In the testing stage, we only track forward to predict the target location.}
	\label{fig:main_figure} 
\end{figure*}

\subsection{Forward-Backward Analysis}
The forward and backward strategy has been investigated in motion analysis scenarios.
Meister et al. \cite{UnFlow} combined the forward-backward consistency estimation and pixel construction to learn optical flows.
Wang et al. \cite{correspondence} leveraged the cycle-consistency across multiple steps temporally to learn feature representations for different tasks.
The differentiable tracker in \cite{correspondence} is deliberately designed to be weak for feature representation learning.
In contrast, aiming at robust visual tracking, we adopt a strong tracking baseline (Siamese correlation filter network), which is not fully-differentiable in the trajectory loop due to the pseudo labeling.
However, by repeating forward tracking and backward verification, we incrementally promote the tracking network by pseudo-labeling based self-training. 
The forward-backward consistency check is also applied in image alignment \cite{aligment_minimization,alignment_3d} and depth estimation \cite{geonet,unsupervised_ego-motion}.
In the visual tracking community, the forward-backward consistency is mainly used for the output reliability or uncertainty measurement. The tracking-learning-detection (TLD) \cite{TLD} uses the Kanade-Lucas-Tomasi (KLT) tracker \cite{KLTtracker} to perform forward-backward matching to detect tracking failures.
Lee et al. \cite{Multihypothesis} proposed to select the reliable base tracker by comparing the geometric similarity, cyclic weight, and appearance consistency between a pair of forward-backward trajectories. However, these methods rely on empirical metrics to identify the target trajectories. In addition, repeatedly performing forward and backward trackings brings in a heavy computational cost for online tracking and largely hurts the real-time performance.
Differently, in TrackingNet \cite{2018trackingnet}, forward-backward analysis is used for evaluating the tracking performance and annotating a sparsely labeled dataset such as Youtube-BoundingBox \cite{youtubeBB} to obtain the per-frame object bounding box labels.
In this work, we target at visual tracking but revisit the forward-backward scheme from a different view, i.e., training a deep visual tracker in under unsupervised manner.

\subsection{Unsupervised Representation Learning}
Our tracking framework relates to unsupervised representation learning. 
Learning feature representations from raw videos under an unsupervised manner has gained increasing attention in recent years.
These approaches typically design ingenious techniques to explore and utilize the free supervision in images or videos.
In \cite{lee2017unsupervised}, the feature representation is learned by shuffling the video frames and then sorting them again to achieve self-supervised training. The multi-layer autoencoder on large-scale unlabeled data has been explored in \cite{le2013building}. Vondrick et al. \cite{vondrick2016anticipating} proposed to anticipate the visual representation of frames in the future.
In \cite{TrackingViaColorizing}, Vondrick et al. colorized gray-scale videos by copying colors from a reference frame to learn a CNN model.
Wang and Gupta \cite{wang2015unsupervised} used the KCF tracker \cite{KCF} to pre-process the raw videos, and then selected a pair of tracked images together with another random patch for learning CNNs using a ranking loss. Our method differs from \cite{wang2015unsupervised} significantly in two aspects. First, we integrate the tracking algorithm into unsupervised training instead of merely utilizing an off-the-shelf tracker as the data pre-processing tool. Second, our unsupervised framework is coupled with a tracking objective function, so the learned feature representation is effective in characterizing the generic target objects.

In the visual tracking community, unsupervised learning has rarely been touched. According to our knowledge, the only related but different approach is the auto-encoder based method \cite{DLT}. However, the encoder-decoder is a general unsupervised framework \cite{olshausen1997sparse}, whereas our unsupervised method is specially designed for the tracking task.
Since the visual objects or scenes in videos typically change smoothly, the motion information of the objects in a forward-backward trajectory loop provides free yet informative self-supervision signals for unsupervised learning, which is naturally suitable for the motion-related visual tracking task.

\section{Proposed Method}\label{sec:proposed approach}

The motivation of our unsupervised learning is shown in Fig.~\ref{fig:main_figure}(a).
We first select a content-rich local region as the target object.
Given this initialized bounding box label, we track forward to predict its location in the subsequent frames.
Then, we reverse the sequence and take the predicted bounding box in the last frame as the pseudo label for backward verification.
The predicted bounding box in the first frame via backward tracking is ideally identical to the original bounding box.
We measure the difference between the forward and backward trajectories using the consistency loss to train the network.
Fig.~\ref{fig:main_figure}(b) shows an overview of our unsupervised Siamese correlation filter network.

In the following, we first revisit the correlation filter as well as the Siamese network. In Section \ref{representation learning}, we present our unsupervised learning prototype for an intuitive understanding. In Section \ref{sec:stable training}, we improve our prototype to facilitate unsupervised training. Finally, training details and online tracking are elaborated in Sections \ref{training details} and \ref{online tracking}, respectively.

\subsection{Revisiting Correlation Tracking}\label{revisit CF}
The Discriminative Correlation Filters (DCFs) \cite{MOSSE,KCF} regress the circularly shifted versions of the input features of a search patch to a soft target response map for target localization.
When training a DCF, we select a template patch $ \bf X $ with the corresponding ground-truth label $ \bf Y $, which is Gaussian-shaped with the peak localized at the target position.
The size of the template patch is usually larger than that of the target.
Fig.~\ref{fig:main_figure} shows an example of the template patch, where there are both target and background contents.
The filter $ \bf W $ can be learned by solving the following ridge regression problem:
\begin{equation}\label{Eq1}
\min_{\bf W}{\|{\bf W}\ast{\bf X}-{\bf Y}\|}^{2}_{2}+\lambda{\|{\bf W}\|}^{2}_{2},
\end{equation}
where $ \lambda $ is a regularization parameter and $ \ast $ denotes the circular convolution. Eq.~\ref{Eq1} can be efficiently calculated in the Fourier domain \cite{MOSSE,DSST,KCF} and the DCF can be computed by
\begin{equation}\label{Eq2}
{\bf W} = \mathscr{F}^{-1}\left( \frac{\mathscr{F}(\bf X)\odot\mathscr{F}^{\star}(Y)}{\mathscr{F}^{\star}(\bf X)\odot\mathscr{F}(X)+\lambda}\right),
\end{equation}
where $\odot$ is the element-wise product, $ \mathscr{F}(\cdot) $ is the Discrete Fourier Transform (DFT), $ \mathscr{F}^{-1}(\cdot) $ is the inverse DFT, and $ \star $ denotes the complex-conjugate operation. In each subsequent frame, given a search patch $ \bf Z $, its corresponding response map $ \bf R $ can be computed in the Fourier domain:
\begin{equation}\label{Eq3}
{\bf R} = {\bf W\ast Z}= {\mathscr F}^{-1} \left( {\mathscr F}^{\star}(\bf W)\odot{\mathscr{F} (\bf Z)} \right).
\end{equation}

The above DCF framework starts from learning the target template's correlation filter (i.e., $ \bf W $) using the template patch and then convolves it with a search patch $\bf Z$ to generate the response. Recently, the Siamese correlation filter network \cite{CFNet,DCFNet} embeds the DCF in the Siamese framework and constructs two shared-weight branches to extract feature representations, as shown in Fig.~\ref{fig:main_figure}(b). The first one is the template branch which takes a template patch $\bf X$ as input and extracts its features to further generate a target template filter via DCF. The second one is the search branch which takes a search patch $\bf Z$ as input for feature extraction. The  template filter is then convolved with the CNN features of the search patch to generate the response map. The advantage of the Siamese DCF network is that both the feature extraction CNN and correlation filter are formulated into an end-to-end framework, so the learned features are more related to the visual tracking scenarios.

\subsection{Unsupervised Learning Prototype}\label{representation learning}

Given two consecutive frames $P_1$ and $P_2$, we crop the template and search patches from them, respectively. By conducting forward tracking and backward verification, the proposed framework does not require additional supervision. The location difference between the initial bounding box and the predicted bounding box in $P_1$ will formulate a consistency loss. We utilize this loss to train the network without ground-truth annotations.

\subsubsection{Forward Tracking}
Following the previous approaches \cite{CFNet,DCFNet}, we build a Siamese correlation filter network to track the initialized bounding box region in frame $P_1$.
After generating the template patch $ \bf T $ from the first frame $P_1$, we compute the corresponding template filter $\bf W_{T}$ as follows:
\begin{equation}\label{Eq4}
{\bf W}_{\bf T} = \mathscr{F}^{-1}\left( \frac{\mathscr{F}(\varphi_{\theta}({\bf T}))\odot\mathscr{F}^{\star}({\bf Y}_{\bf T})}{\mathscr{F}^{\star}(\varphi_{\theta}({\bf T}))\odot\mathscr{F}(\varphi_{\theta}({\bf T}))+\lambda} \right),
\end{equation}
where $ \varphi_{\theta}(\cdot) $ denotes the CNN feature extraction operation with trainable network parameters $ \theta $, and $ \bf Y_{T} $ is the label of the template patch $ \bf T $. This label is a Gaussian response centered at the initialized bounding box center.
Once we obtain the learned template filter ${\bf W}_{\bf T}$, the response map of a search patch $ \bf S $ from frame $P_2$ can be computed by
\begin{equation}\label{Eq5}
{\bf R}_{\bf S} = \mathscr{F}^{-1}(\mathscr{F}^{\star}({\bf W}_{\bf T})\odot\mathscr{F}(\varphi_{\theta}({\bf S}))).
\end{equation}
If the ground-truth Gaussian label of patch $ \bf S $ is available, the network $ \varphi_{\theta}(\cdot) $ can be trained by computing the $ L_{2} $ distance between $ \bf R_{S} $ and the ground-truth label. Different from the supervised framework, in the following, we present how to train the network without requiring labels by exploiting backward trajectory verification.

\subsubsection{Backward Tracking}
After generating the response map ${\bf R}_{\bf S}$ for frame $P_2$, we create a pseudo Gaussian label centered at its maximum value, which is denoted by $ {\bf Y}_{\bf S} $.
In backward tracking, we switch the role between the search patch and the template patch.
By treating $\bf S$ as the template patch, we generate a template filter $\bf W_{S}$ using the pseudo label ${\bf Y}_{\bf S}$.
The template filter $\bf W_{S}$ can be learned using Eq.~\ref{Eq4} by replacing $\bf T$ with $\bf S$ and replacing ${\bf Y}_{\bf T}$ with ${\bf Y}_{\bf S}$, as follows:
\begin{equation}\label{new_eq1}
	{\bf W}_{\bf S} = \mathscr{F}^{-1}\left( \frac{\mathscr{F}(\varphi_{\theta}({\bf S}))\odot\mathscr{F}^{\star}({\bf Y}_{\bf S})}{\mathscr{F}^{\star}(\varphi_{\theta}({\bf S}))\odot\mathscr{F}(\varphi_{\theta}({\bf S}))+\lambda} \right).
\end{equation}
Then, we generate the response map ${\bf R}_{\bf T}$ of the template patch through Eq.~\ref{Eq5} by replacing ${\bf W}_{\bf T}$ with $\bf W_{S}$ and replacing $\bf S$ with $\bf T$, as shown in Eq.~\ref{new_eq2}.
\begin{equation}\label{new_eq2}
	{\bf R}_{\bf T} = \mathscr{F}^{-1}(\mathscr{F}^{\star}({\bf W}_{\bf S})\odot\mathscr{F}(\varphi_{\theta}({\bf T}))).
\end{equation}
Note that we only use one Siamese correlation filter network for executing forward and backward trackings. The network parameters $ \theta $ are fixed during the tracking steps.

\subsubsection{Consistency Loss Computation}
After forward and backward tracking, we obtain the response map ${\bf R}_{\bf T}$. Ideally, ${\bf R}_{\bf T}$ should be a Gaussian label with the peak located at the initialized target position. In other words,
$ \bf R_{T} $ should be as similar as the originally given label $ \bf Y_{T} $.
Therefore, the representation network $ \varphi_{\theta}(\cdot) $ can be trained under an unsupervised manner by minimizing the reconstruction error as follows:
\begin{equation}\label{Eq6}
{\cal L}_{\text{un}} = \|{\bf R}_{\bf T}-{\bf Y}_{\bf T}\|^{2}_{2}.
\end{equation}

\begin{figure}
	\centering
	\includegraphics[width=7.2cm]{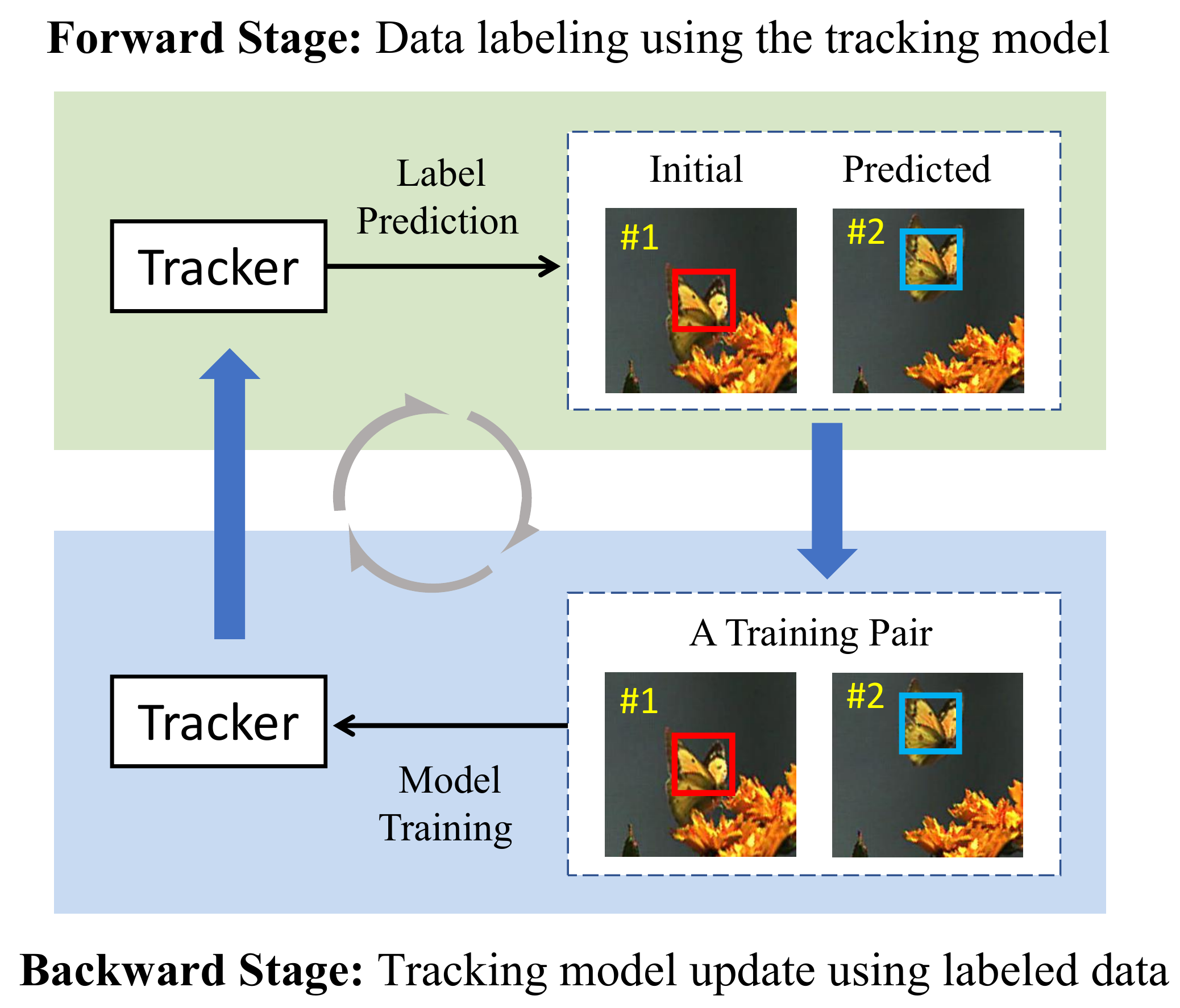}
	\caption{The intuition of pseudo-labeling based self-training. We use the same network for both forward and backward predictions. The forward stage generates a pseudo label for the search patch. The backward stage updates the tracking network using training pairs via loss back-propagation. During training iterations, the response map of the template gradually approaches the initial label via self supervision.}
	\label{fig:iteration} 
\end{figure}

Our unsupervised learning can be viewed as an incremental self-training process that iteratively predicts labels and updates the model to steadily improve the tracking capability. 
Fig.~\ref{fig:iteration} shows the intuition, where we use the same network for both forward and backward predictions.
In the forward tracking, we generate a pseudo label $ {\bf Y}_{\bf S} $ for the search patch $ \bf S $.
Then we treat generated ${\bf Y}_{\bf S}$ as the label of $ \bf S $ and create a corresponding sample. 
Using these labeled training pairs (i.e., with initial or pseudo labels), we can update the Siamese correlation filter network in a similar way to supervised learning.
%
%
%
During loss back-propagation, we follow the Siamese correlation filter methods \cite{DCFNet,SACF} to update the network:
\begin{equation}\label{Eq7}
\begin{aligned}
\frac{\partial {\cal L}_{\text{un}}}{\partial \varphi_{\theta}(\bf T)} &= \mathscr{F}^{-1}\left( \frac{\partial {\cal L}_{\text{un}}}{\partial \left(\mathscr{F}\left(\varphi_{\theta}(\bf T)\right)\right)^{\star}} + \left( \frac{\partial {\cal L}_{\text{un}}}{\partial \left(\mathscr{F}\left(\varphi_{\theta}(\bf T)\right)\right)} \right)^{\star} \right),\\
\frac{\partial {\cal L}_{\text{un}}}{\partial \varphi_{\theta}(\bf S)} &= \mathscr{F}^{-1}\left( \frac{\partial {\cal L}_{\text{un}}}{\partial \left(\mathscr{F}\left(\varphi_{\theta}({\bf S})\right)\right)^{\star}}\right).
\end{aligned}
\end{equation}


The above unsupervised training process is based on the forward-backward consistency between two frames, which is summarized by Algorithm \ref{code}. In the next section, we extend this prototype framework to consider multiple frames for better network training.

\subsection{Enhancement for Unsupervised Learning}\label{sec:stable training}
The proposed unsupervised learning method constructs the objective function based on the consistency between ${\bf R}_{\bf T}$ and $\bf Y_{T}$. In practice, the tracker may deviate from the target in the forward tracking but still return to the original position during the backward process.
However, the proposed loss function does not penalize this deviation because of the consistent trajectories.
Meanwhile, the raw videos may contain textureless or occluded training samples that deteriorate the unsupervised learning process.
In this section, we propose a multi-frame validation scheme and a cost-sensitive loss to tackle these two limitations.

\subsubsection{Multi-frame Validation} \label{Multiple Frames Validation}
We propose a multi-frame validation approach to enlarge the trajectory inconsistency when the tracker loses the target. Our intuition is to incorporate more frames during training to reduce the limitation that the erroneous localization in the subsequent frame successfully backtraces to the initial position in the first frame. In this way, the reconstruction error in Eq.~\ref{Eq6} will effectively capture the inconsistent trajectory. 
As shown in Fig.~\ref{fig:iteration}, adding more frames in the forward stage further challenges the model tracking capability.

\begin{algorithm}[t] \label{code}
	\caption{Unsupervised training prototype}
	\LinesNumbered
	\KwIn{Unlabeled videos.}
	\KwOut{Pretrained tracking network $ \varphi_{\theta}(\cdot) $.}
	Crop the patches (i.e., $ \bf T $ and $ \bf S $) from the raw videos;\\
	Initialize the CNN model $ \varphi_{\theta}(\cdot) $ with random weights $ \theta $;\\
	\For{$ \text{each training epoch} $}{
		\For{$\text{each training pair} $}{
			Obtain $ \varphi_{\theta}({\bf T})$ and $ \varphi_{\theta}({\bf S}) $;\\
			//{\tt { \color{blue} Forward Trajectory}}\\
			Construct $ \bf W_{T} $ using $ \varphi_{\theta}({\bf T}) $ and $ \bf Y_{T} $ (Eq.~4);\\
			Compute $ {\bf R}_{{\bf S}} $ using $ \bf W_{T} $ (Eq.~5) and obtain the pseudo label of $ {\bf S} $;\\
			//{\tt { \color{blue} Backward Trajectory}}\\
			Construct $ {\bf W}_{{\bf S}} $ using $ \varphi_{\theta}({\bf S}) $ and $ {\bf Y}_{{\bf S}} $ (Eq.~6);\\
			Compute the response map $ {\bf R}_{\bf T} $ of $ \bf T $ (Eq.~7);\\
			//{\tt { \color{blue} Calculate Consistency Loss}}\\
			Compute the consistency loss of $ {\bf Y}_{\bf T} $ and $ {\bf R}_{\bf T} $ (Eq.~8);\\
		}
		Update network $ \varphi_{\theta}(\cdot) $ using the computed loss;\\
	} 
\end{algorithm}

\begin{figure}
	\centering
	\includegraphics[width=8.2cm]{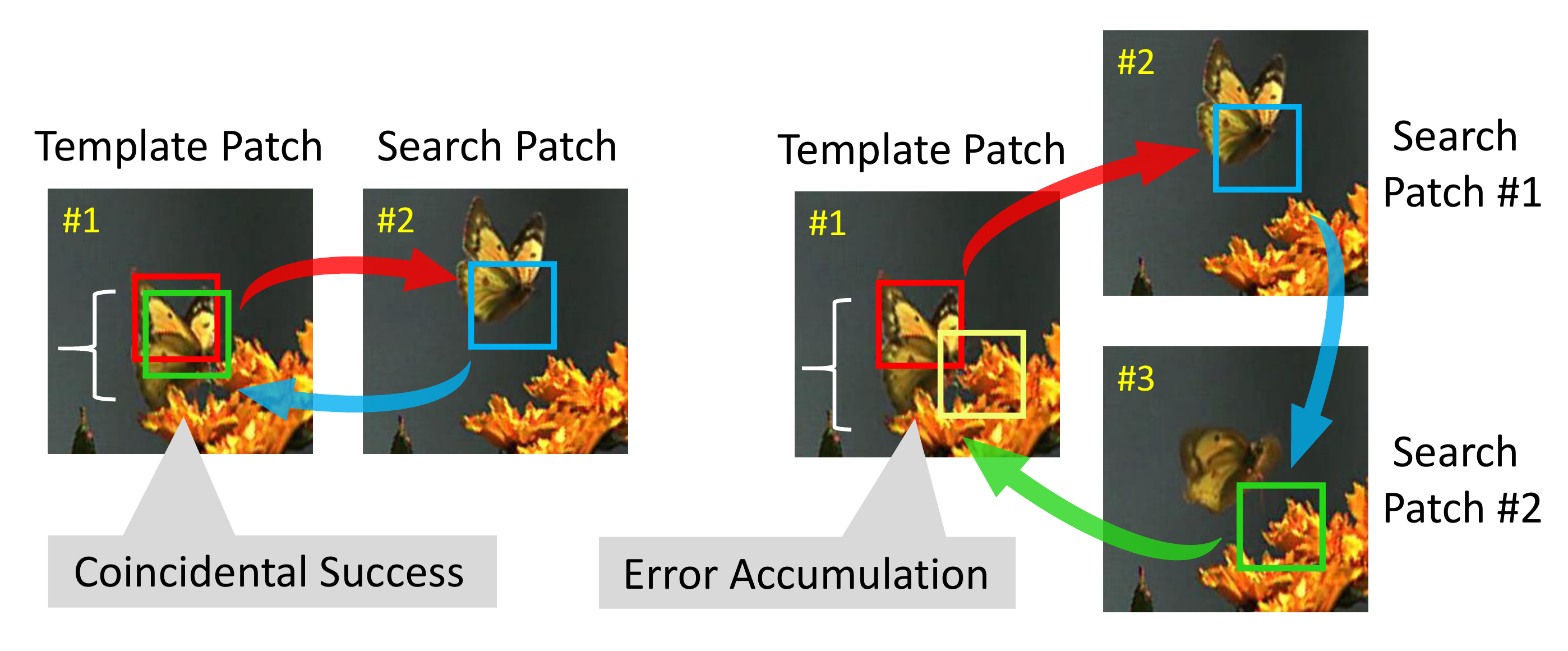}
	\caption{Single frame validation and multi-frame validation. The inaccurate localization in single frame validation may not be captured as shown on the left. By involving more frames as shown on the right, we accumulate the localization errors to break the prediction consistency during forward and backward trackings.}
	\label{fig:multi-frame validation} 
\end{figure}

Our unsupervised learning prototype can be easily extended to multiple frames. To build a trajectory cycle using three frames, we can involve another frame $P_3$ which is the subsequent frame after $P_2$. We crop a search patch ${\bf S}_{1}$ from $P_2$ and another search patch ${\bf S}_2$ from $P_3$. If the generated response map $ {\bf R}_{{\bf S}_1}$ is different from its corresponding ground-truth response, the difference tends to become larger in the next frame $P_3$.
As a result, the inconsistency is more likely to appear in backward tracking, and the generated response map $\bf R_{T}$ is more likely to differ from $\bf Y_{T}$, as shown in Fig.~\ref{fig:multi-frame validation}.
By involving more search patches during forward and backward trackings, the proposed consistency loss will be more effective to penalize the inaccurate localizations.

We can further extend the number of frames utilized for multi-frame validation.
The length of trajectory will increase as shown in Fig.~\ref{fig:trajectory length}.
The limitation of consistent trajectory when losing the target is more unlikely to affect the training process.
Let $ {\bf R}_{({\bf S}_{k}\to{\bf T})} $ denote the response map of the template $ \bf T $, which is generated (or tracked) by the DCF trained using the $ k $-th search patch $ {\bf S}_k $. The corresponding consistency loss function can be computed as follows:
\begin{equation}\label{Eq8}
{\cal L}_{k} = \left\|{\bf R}_{({\bf S}_{k}\to{\bf T})}-{\bf Y}_{\bf T}\right\|^{2}_{2}.
\end{equation}
Considering different trajectory cycles, the multi-frame consistency loss can be computed by
\begin{equation}\label{Eq9}
{\cal L}_{\text{un}} = \sum_{k=1}^{M}{\cal L}_{k},
\end{equation}
where $ k $ is the index of the search parch. Taking Fig.~\ref{fig:trajectory length}(c) as an example, the final consistency objective contains three losses (i.e., $ M=3 $ in Eq.~\ref{Eq9}), which are denoted by the blue, green, and red cycles in Fig.~\ref{fig:trajectory length}(c), respectively.

\subsubsection{Cost-sensitive Loss}\label{cost_sensitive loss}
\begin{figure}
	\centering
	\includegraphics[width=6.0cm]{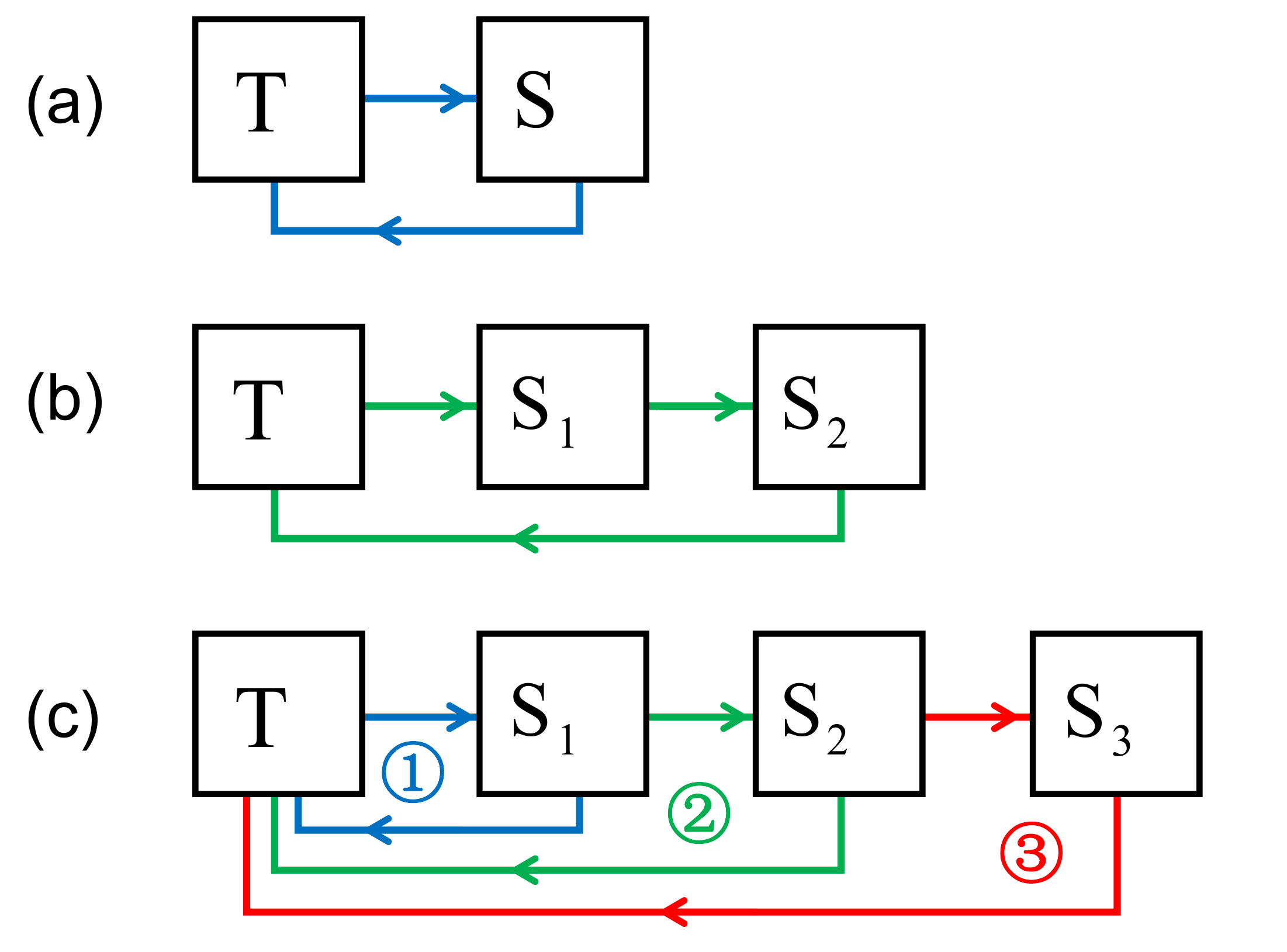}
	\caption{An overview of multi-frame trajectory consistency. We denote $ \bf T $ as a template and $ \bf S $ as a search patch, respectively. Our unsupervised training prototype is shown in (a), where only two frames are involved. Using more frames as shown in (b) and (c), we can gradually improve the training performance to overcome consistent trajectories when losing the target.}
	\label{fig:trajectory length} 
\end{figure}

We initialize a bounding box region as a training sample in the first frame during unsupervised training.
The image content within this bounding box region may contain arbitrary or partial objects.
Fig.~\ref{fig:crop_example} shows an overview of these regions.
To alleviate the background interference, we propose a cost-sensitive loss to effectively exclude noisy samples for network training.
For simplicity, we use three consecutive frames as an example to illustrate sample selection, which can be naturally extended to more frames.
The pipeline of using three frames is shown in Fig.~\ref{fig:trajectory length}(b).

During unsupervised learning, we construct multiple training triples from video sequences. For a trajectory containing three frames, each training triple consists of one initialized template patch $\bf T$ in frame $P_1$ and two search patches ${\bf S}_{1}$ and ${\bf S}_{2}$ in the subsequent frames $P_2$ and $P_3$, respectively. We use several triples to form a training batch for Siamese network learning.
In practice, we find that some training triples with extremely high losses prevent network training from convergence.
To reduce these outlier effects in pseudo-labeling based self-training, we exclude 10\% of the whole training triples which contain the highest loss values. Their losses can be computed using Eq.~\ref{Eq8}.
To this end, we assign a binary weight ${\bf A}^{i}_{\text{drop}}$ to each training triple.
All these weights constitute a vector $\bf A_{\text{drop}}$, where 10\% of its elements are 0 and the others are 1.

In addition to the outlier training pairs, the raw videos include meaningless image patches, where there are textureless backgrounds or still objects. In these patches, the objects (e.g., sky, grass, or tree) do not contain big movements.
We assign a motion weight vector $\bf A_{\text{motion}}$ to all the training pairs to increase the large motion effect for network learning.
Each element ${\bf A}_{\text{motion}}^i$ within this vector can be computed by
\begin{equation}\label{Eq10}
{\bf A}_{\text{motion}}^i = \left\|{\bf R}_{{\bf S}_{1}}^i-{\bf Y}_{\bf T}^i\right\|^{2}_{2} +\left\|{\bf R}_{{\bf S}_{2}}^i-{\bf Y}_{{\bf S}_{1}}^i\right\|^{2}_{2},
\end{equation}
where $ {\bf R}_{{\bf S}_{1}}^i $ and $ {\bf R}_{{\bf S}_{2}}^i $ are the response maps in the $i$-th training pair, and $ {\bf Y}_{{\bf T}}^i $ and $ {\bf Y}_{{\bf S}_{1}}^i $ are the corresponding initial (or pseudo) labels. Eq.~\ref{Eq10} calculates the target motion difference from frame $P_1$ to $P_2$ and $P_2$ to $P_3$. When the value of ${\bf A}^{i}_{\text{motion}}$ is large, the target object undergoes fast motion in this trajectory. On the other hand, the large value of ${\bf A}^{i}_{\text{motion}}$ represents the hard training pair which the network should pay more attention to.
We normalize the motion weight and the binary weight as follows:
\begin{equation}\label{Eq11}
{\bf A}_{\text{norm}}^{i} = \frac{{\bf A}^{i}_{\text{drop}}\cdot{\bf A}^{i}_{\text{motion}}}{\sum_{i=1}^{N}{\bf A}^{i}_{\text{drop}}\cdot{\bf A}^{i}_{\text{motion}}},
\end{equation}
where $ N $ is number of the training pairs in a mini-batch. The sample weight $ {\bf A}_{\text{norm}}^{i} $ serves as a scalar that reweighs the training data without gradient back-propagation.

The final unsupervised loss for the case of Fig.~\ref{fig:trajectory length}(b) in a mini-batch is computed as:
\begin{equation}\label{Eq12}
{\cal L}_{\text{3-frame}} = \frac{1}{N}\sum_{i=1}^{N}{\bf A}^{i}_{\text{norm}}\cdot \left\|{\bf R}^{i}_{({\bf S}_{2}\to{\bf T})}-{\bf Y}^{i}_{\bf T}\right\|^{2}_{2}.
\end{equation}
We can naturally extend Eq.~\ref{Eq12} to the following by using more frames to construct trajectories of different lengths, as illustrated by the toy example of Fig.~\ref{fig:trajectory length}(c). Combining with Eq.~\ref{Eq9}, we compute the final unsupervised loss function using $M$ subsequent frames as:
\begin{equation}\label{Eq13}
{\cal L}_{\text{final}} = \frac{1}{N}\sum_{k=1}^{M}\sum_{i=1}^{N}{\bf A}^{i}_{\text{norm}}\cdot{\cal L}^{i}_{k},
\end{equation}
where $ {\cal L}^{i}_{k} = \left\|{\bf R}^{i}_{({\bf S}_{k}\to{\bf T})}-{\bf Y}^{i}_{\bf T}\right\|^{2}_{2} $ is similar to that in Eq.~\ref{Eq8} but with the index $ i $ for differentiating different samples in a mini-batch.

\subsection{Unsupervised Training Details} \label{training details}
{\flushleft \bf Network Structure.} We follow the DCFNet \cite{DCFNet} to use a shallow Siamese network consisting of two convolutional layers for tracking. This shallow structure is demonstrated effective in CFNet \cite{CFNet} to integrate the DCF formulation. The filter sizes of these convolutional layers are $3\times3\times3\times32$ and $3\times3\times32\times32$, respectively. Besides, a local response normalization (LRN) layer is employed at the end of convolutional layers following \cite{DCFNet}. This lightweight structure enables efficient forward inferences for online tracking.

{\flushleft \bf Training Data.} We choose ILSVRC 2015 \cite{ILSVRC2015} as our training data, which is the same dataset employed by existing supervised trackers. In the data pre-processing step, supervised approaches \cite{SiamFc,CFNet,DCFNet} require per-frame labels. Besides, the frames will be removed, where the target object is occluded, partially out-of-view, or in an irregular shape (e.g., snake). The data pre-precessing for the supervised approaches is time-consuming with human labor. In contrast, our method does not rely on manually annotated labels for data pre-processing.

\begin{figure}[t]
	\centering
	\includegraphics[width=8.2cm]{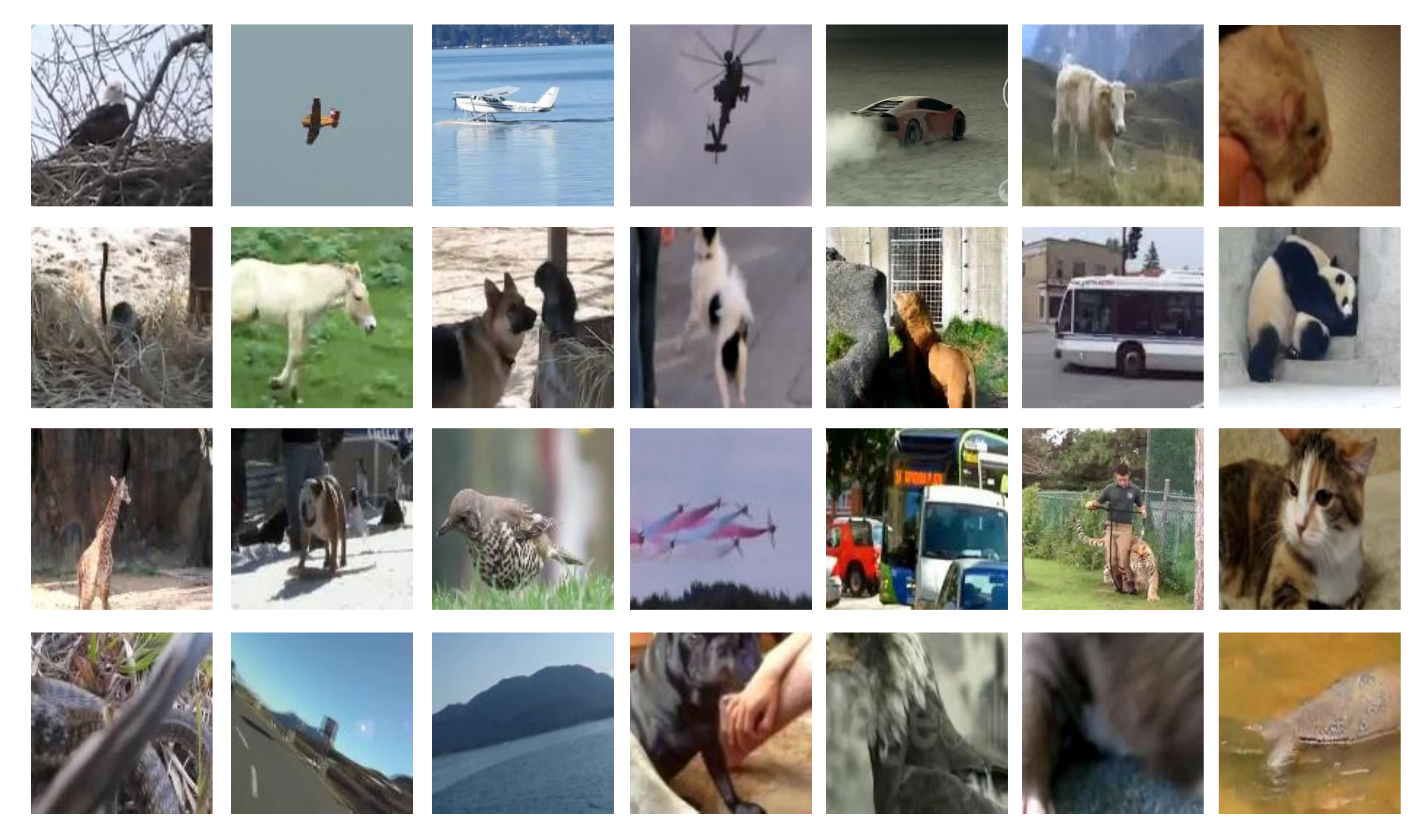}
	\caption{Examples of the cropped image patches from ILSVRC 2015 \cite{ILSVRC2015}. Most of these samples contain meaningful objects, while some samples are less meaningful (e.g., last row).}
	\label{fig:crop_example} 
\end{figure}

\begin{figure}
	\centering
	\includegraphics[width=8.2cm]{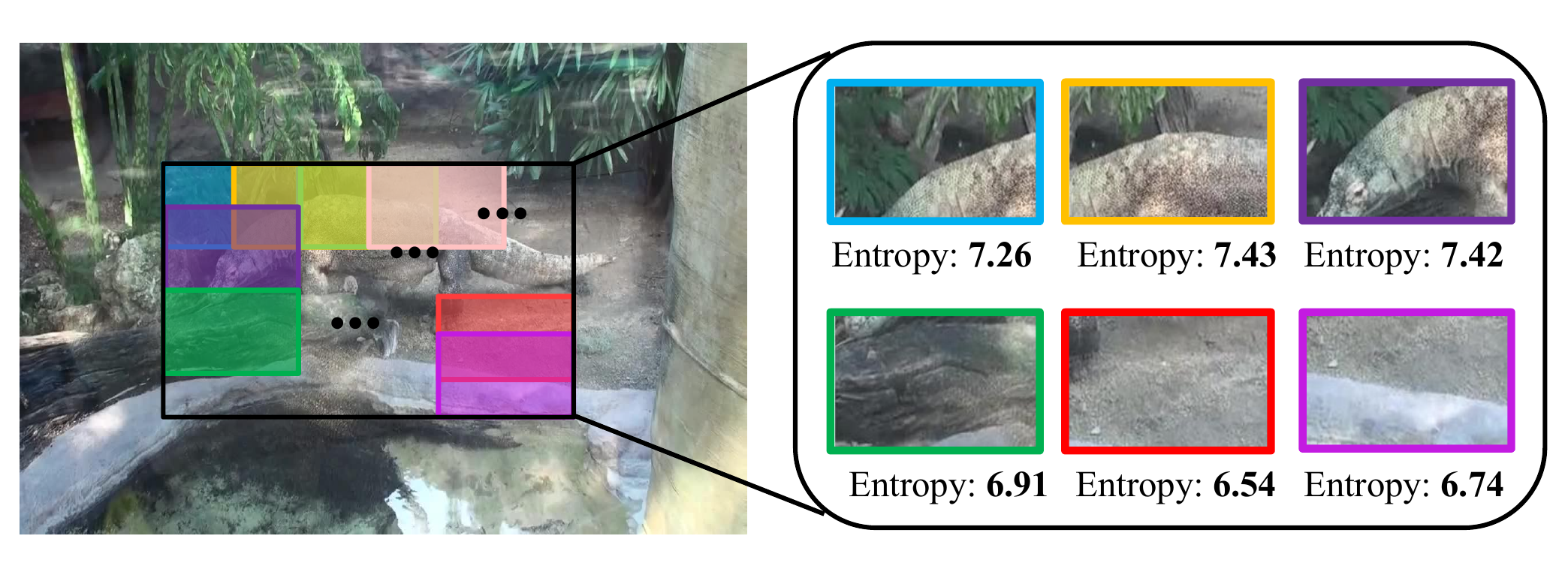}
	\caption{The illustration of training samples generation. The proposed method crops $ 5\times5 $ patch candidates in the center region of the initial frame. Then we select the image patch with the highest image entropy. As shown in the right figure, the background patches (e.g., labeled by green and red boxes) have a small image entropy.}
	\label{fig:image_entropy} 
\end{figure}

In our approach, for the first frame in a raw video, we crop overlapped small patches ($ 5\times5 $ in total) by sliding windows as shown in Fig.~\ref{fig:image_entropy}. Then, we compute the image entropy of each image patch. Image entropy effectively measures the content variance of an image patch. When an image patch only contains the unitary texture (e.g., the sky), the entropy of this patch approaches 0. When an image patch contains textured content, the entropy will become higher.
We select the cropped image patch containing the highest image entropy.
This image patch initializes the KCF \cite{KCF} tracker for localization in the subsequent frames.
Then, we crop a larger image patch with a padding of 2 times of the target size following DCFNet \cite{DCFNet}, which is further resized to $125\times125$ as the input of our network.
Fig.~\ref{fig:crop_example} exhibits some examples of the cropped patches.
We randomly choose 4 cropped patches from the continuous 10 frames in a video to form a training trajectory, and one of them is defined as the template and the rest as search patches. This is based on the assumption that the center located target objects are unlikely to move out of the cropped region in a short span of time.
We track the content in the image patch regardless of specific object categories.
Although this entropy-based method may not accurately select a target region and the KCF tracker is not robust enough to track the cropped region, this method can well alleviate the meaningless background regions.

%

\subsection{Online Object Tracking}\label{online tracking}
After offline unsupervised learning, we perform online tracking in the way of forward tracking as illustrated in Section \ref{representation learning}. We online update DCF to adapt to the target appearance changes. The DCF update follows a moving average operation shown as follows:
\begin{equation}\label{Eq14}
{\bf W}_{t} = (1-\alpha_{t}){\bf W}_{t-1}+\alpha_{t}{\bf W},
\end{equation}
where $ \alpha_{t}\in[0,1] $ is the linear interpolation coefficient. The target scale is estimated through a patch pyramid with scale factors $ \{ a^{s}|a=1.015, s=\{-1, 0, 1\}\} $ \cite{SRDCF}. We name our Tracker as LUDT (i.e., Learning Unsupervised Deep Tracking).
Besides, we update our model adaptively via $ \alpha_{t} $ and follow the superior DCF formulation as that in ECO \cite{ECO}. We name the improved tracker as LUDT+.

We keep the notation of our preliminary tracker UDT and UDT+ \cite{UDT} in the following experiment section. Our previous UDT uses a 3-frame cycle (Fig.~\ref{fig:trajectory length}(b)) and simply crops the center patch in raw videos. LUDT improves UDT in two aspects: (1) LUDT combines different trajectory cycles as shown in Fig.~\ref{fig:trajectory length}(c), and (2) LUDT utilizes image entropy to select the informative image patches instead of the center crop. LUDT+ and UDT+ improve LUDT and UDT by adopting some online tracking techniques (e.g., adaptive update) proposed in \cite{ECO}, respectively.


\section{Experiments}\label{experiment}

In this section, we first analyze the effectiveness of our unsupervised training framework and discuss our network potentials. Then, we compare our tracker LUDT against state-of-the-art trackers on both standard and recently released large-scale benchmarks including OTB-2013 \cite{OTB-2013}, OTB-2015 \cite{OTB-2015}, Temple-Color \cite{TempleColor128}, VOT2016 \cite{VOT2016}, VOT2017/2018 \cite{VOT2018}, LaSOT \cite{LaSOT}, and TrackingNet \cite{2018trackingnet}.

\subsection{Experimental Details} \label{sec:Setup}
In our experiments, we use the stochastic gradient descent (SGD) with a momentum of 0.9 and a weight decay of 0.005 to train our model. Our unsupervised network is trained for 50 epochs with a learning rate exponentially decaying from $ 10^{-2} $ to $ 10^{-5} $ and a mini-batch size of 32. We set the trajectory length as 4. All the experiments are executed on a PC with 4.00GHz Intel Core I7-4790K and NVIDIA GTX 1080Ti GPU. 
On a single GPU, our LUDT and LUDT+ exhibit about 70 FPS and 55 FPS, respectively\footnote{The source code is provided at \url{https://github.com/594422814/UDT}}.

The proposed method is evaluated on seven benchmarks. On the OTB-2013/2015, TempleColor, LaSOT, and TrackingNet datasets, we use one-pass evaluation (OPE) with distance and overlap precision metrics. The distance precision threshold is set as 20 pixels. The overlap success plot uses thresholds ranging from 0 to 1, and the area-under-curve (AUC) is computed to evaluate the overall performance. On the VOT2016 and VOT2017/2018 datasets, we measure the performance using Expected Average Overlap (EAO).

\subsection{Ablation Experiments and Discussions}\label{ablation}
\subsubsection{Improvements upon UDT}
Our preliminary tracker UDT \cite{UDT} adopts a three-frame validation (i.e., Fig.~\ref{fig:trajectory length}(b)) and the center crop for sample generation. 
The improvement upon UDT is that we construct a multi-supervision consistency loss function using more frames. We denote this strategy as Trajectory Enlargement (TE) in Table \ref{table:improvements on UDT}. Meanwhile, we select the RoI (region of interest) from raw videos using the image entropy and KCF tracker, while only the center region is utilized in UDT. We denote RoI Selection as RS in the table. Note that the performance of UDT has been close to that of its supervised configuration and exceeded several supervised trackers. Moreover, under the same training configuration, LUDT steadily improves UDT by using TE and RS during training. LUDT achieves 60.2\% and 51.5\% in AUC on the OTB-2015 and Temple-Color benchmarks, respectively.

\setlength{\tabcolsep}{2pt}
\begin{table}
	\scriptsize
	\begin{center}
		\caption{Ablation study of Trajectory Enlargement (TE) and RoI Selection (RS). We denote UDT as our preliminary tracker \cite{UDT}. We integrate TE and RS into UDT during training and report the performance improvement. The evaluation metrics are DP and AUC scores on the OTB-2015 and Temple-Color datasets.} \label{table:improvements on UDT}
		\begin{tabular*}{5.3 cm}{@{\extracolsep{\fill}}|l|c|c|} 
			\hline
			~ &OTB-2015 & Temple-Color   \\
			~ &DP~/~AUC (\%) & DP~/~AUC (\%)   \\
			\hline
			\hline
			~UDT \cite{UDT} & 76.0~/~59.4 &65.8~/~50.7 \\
			~UDT + TE & 76.5~/~59.8 &66.7~/~51.2 \\
			~UDT + RS &76.5~/~60.0 &66.9~/~51.3 \\
			~UDT + TE + RS &76.9~/~60.2 &67.1~/~51.5 \\
			\hline
		\end{tabular*}
	\end{center}
\end{table}

\setlength{\tabcolsep}{2pt}
\begin{table}
	\scriptsize
	\begin{center}
		\caption{Comparison results of the DCFNet tracking framework with different feature extractors. Random: the randomly initialized feature extractor without pre-training. HOG: adopting HOG \cite{HOG} without deep features. ED: the backbone network trained via encoder-decoder \cite{DLT}. The evaluation metrics are DP and AUC scores on OTB-2015.} \label{table:baselines}
		\begin{tabular*}{5.8 cm}{@{\extracolsep{\fill}}|l|cccc|} 
			\hline
			~ &Random &HOG &ED & LUDT (Ours)   \\
			\hline
			\hline
			~DP (\%) & 59.1 &69.2 &71.6 &76.9\\
			~AUC (\%) & 46.9 &52.1 &54.5  &60.2\\
			\hline
		\end{tabular*}
	\end{center}
\end{table}

\subsubsection{Baseline Performance}\label{ablation:baseline performance}
To verify the effectiveness of the proposed unsupervised framework, we evaluate our tracker using different feature extractors.
As shown in Table \ref{table:baselines}, without pre-training, the model still exhibits a weak tracking capability, which can be attributed to the discriminating power of the correlation filter. By adopting the empirical HOG representations, the performance is still significantly lower than ours. Furthermore, we leverage the auto-encoder framework \cite{DLT} to train the backbone network under an unsupervised manner using the same training data.
From Table~\ref{table:baselines}, we can observe that our approach is superior to the encoder-decoder in this tracking scenario since our forward-backward based unsupervised training is tightly related to object tracking.

\subsubsection{Training Data}
We evaluate the tracking performance using different data pre-processing strategies. The results are shown in Table \ref{table:data pre-processing}. Our unsupervised LUDT method uses the last RoI selection strategy. During the evaluation, we keep the remaining modules fixed in LUDT.

{\noindent \bf Comparison with Full Supervision}. Using the same videos (i.e., ILSVRC 2015 \cite{ILSVRC2015}), we conduct the supervised training of our network. The supervised learning with ground-truth annotations can be regarded as the upper bound of our unsupervised learning. We observe that the performance gap is small (i.e., 2.6\% AUC) between the trackers trained using unsupervised learning (60.0\% AUC) and fully supervised learning (62.6\% AUC).

\setlength{\tabcolsep}{2pt}
\begin{table}
	\scriptsize
	\begin{center}
		\caption{Evaluation results of our network trained using different data pre-processing strategies. Our LUDT tracker uses RoI selection via image entropy for unsupervised training. The evaluation metrics are DP and AUC scores on the OTB-2015 dataset.} \label{table:data pre-processing}
		\begin{tabular*}{8.2 cm} {@{\extracolsep{\fill}}|l|cccc|}
			\hline
			~ & Groundtruth & Groundtruth label  &Center & RoI Selection      \\
			~ & label & with deviations &cropping  & via entropy    \\
			\hline
			& Full supervision & Weak supervision & Unsupervision & Unsupervision \\
			\hline
			\hline
			~DP (\%) &80.6 &78.9 &76.0  &76.5     \\
			~AUC (\%) &62.6 &61.4 &59.4  &60.0   \\
			\hline
		\end{tabular*}
	\end{center}
\end{table}

{\noindent \bf Comparison with Weak Supervision}. On ILSVRC 2015, we add deviations to the ground-truth boxes to crop the training samples. The deviations range from -20 pixels to 20 pixels randomly. The reason for setting sample deviations from the ground-truth bounding boxes is that we aim to simulate the inaccurate object localizations on in-the-wild videos using existing object detection or optical flow approaches. We assume that these deviated samples are predicted by existing methods and then utilized to train our unsupervised network. In Table~\ref{table:data pre-processing}, we observe that our tracker learned by these weakly labeled samples is comparable with the supervised results (61.4\% vs. 62.6\% AUC). Note that 20 pixels deviations can be achieved with many object localization methods. The comparable performance indicates that our method can be applied to raw videos with weakly or sparsely labeled annotations (e.g., the dataset Youtube-BB \cite{youtubeBB}). On the other hand, existing object detectors and models are mostly trained by supervised learning. To ensure our method to be fully unsupervised, we use two unsupervised data pre-processing methods: center cropping and RoI selection based on entropy.

{\noindent \bf Center Cropping}. In center cropping, we crop the center region of the video frame. Although we crop a fixed region of the image, the image content appears randomly in this region and we denote this operation as center cropping. There may be meaningless content (e.g., textureless sky and ocean) in this region to disturb our unsupervised learning. The tracker learned by center cropping achieves an AUC score of 59.4\%.

{\noindent \bf RoI Selection}. We use the entropy-based image patch selection as illustrated in Section \ref{training details}. Compared to the center cropping, image-entropy based selection can suppress the meaningless background samples such as sky and grass, and the KCF tracker is able to capture the selected informative region in the subsequent frames. The RoI selection achieves better performance than center cropping with an AUC score of 60.0\%.

\setlength{\tabcolsep}{1.5pt}
\begin{table}
	\scriptsize
	\begin{center}
		\caption{Evaluation results of our unsupervised model trained using different trajectory lengths. Note that the 4 and 5 frames validations conduct multiple self-supervisions as illustrated in Fig.~\ref{fig:trajectory length}. The evaluation metrics are DP and AUC scores on the OTB-2015 dataset. Compared with 3 frames validation, using more frames further improves the tracking accuracy.} \label{table:trajectory length}
		\begin{tabular*}{7.8 cm} {@{\extracolsep{\fill}}|l|cccc|}
			\hline
			~Frame Number & 2 frames & 3 frames  & 4 frames & 5 frames  \\
			& Fig. \ref{fig:trajectory length}(a) & Fig. \ref{fig:trajectory length}(b)  & Fig. \ref{fig:trajectory length}(c) & akin to Fig. \ref{fig:trajectory length}(c)  \\
			\hline
			\hline
			~DP (\%) &73.2  &76.0  &76.8  &76.8 \\
			~AUC (\%)  &57.4 &59.4 &59.8  &59.7 \\
			\hline
		\end{tabular*}
	\end{center}
\end{table}

\subsubsection{Trajectory Length}

As discussed in Section \ref{Multiple Frames Validation}, trajectory enlargement helps measure the consistency loss when the tracker loses RoI. In Table \ref{table:trajectory length}, we show the performance with different trajectory lengths on the OTB-2015 dataset. We use center cropping to generate training samples following UDT for comparison. The prototype of our unsupervised learning is denoted as 2 frames validation. By incorporating the third frame, the learned tracker achieves improvement (i.e., 2.8\% DP and 2.0\% AUC). The 4 frames validation proposed in this work not only extends the trajectory length but also combines multiple self-supervision constraints, which further improves the accuracy. However, the 5 frames validation seems to be less effective. It may be because the validation with 4 frames already contains adequate self-supervision and effectively measures the consistency loss.

\subsubsection{Cost-sensitive Loss}
On the OTB-2015 dataset, without hard sample reweighing (i.e., ${\bf A}_{\text{motion}}$ in Eq.~\ref{Eq11}), the performance of our LUDT tracker drops about 1.5\% DP and 1\% AUC scores. We did not conduct the ablation study of the sample dropout because we observe that the unsupervised training cannot well converge without $ {\bf A}_{\text{drop}} $ illustrated in Eq.~\ref{Eq12}.

\setlength{\tabcolsep}{2pt}
\begin{table}
	\scriptsize
	\begin{center}
		\caption{A performance study by adding additional training data. Adding more unlabeled data steadily improves the tracking results. The evaluation metrics are DP and AUC scores on the OTB-2015 dataset.} \label{table:more training data}
		\begin{tabular*}{7.8 cm} {@{\extracolsep{\fill}}|l|cccc|}
			\hline
			~ & LUDT & Few-shot fine-tune & More data  & More data  \\
			~ & &OTB-2015 & OxUvA & LaSOT  \\
			\hline
			\hline
			~DP (\%) &76.9 &78.1 & 77.6 & 78.2\\
			~AUC (\%)  &60.2 &61.5 & 61.4 &62.0 \\
			\hline
		\end{tabular*}
	\end{center}
\end{table}

\subsubsection{Unlabled Data Augmentation}

{\noindent \bf Few-shot Domain Adaptation}. To better fit a new domain such as OTB, we construct a small training set by collecting the first several frames (e.g., 5 frames in our experiment) from the videos in OTB-2015 with only the ground-truth bounding box in the first frame available. Using these limited samples, we fine-tune our network by 100 iterations using the forward-backward pipeline, which takes about 6 minutes. As our learning method is unsupervised, we can utilize the frames from test sequences to adapt our tracker. Table~\ref{table:more training data} shows that performance is further improved by using this strategy. Our offline unsupervised training learns general feature representation, which can be transferred to an interested domain (e.g., OTB videos) using few-shot domain adaptation. This domain adaptation is similar to that in MDNet \cite{MDNet}, while our network parameters are initially offline learned in an unsupervised manner.

{\noindent \bf Additional Internet Videos}. We also utilize more unlabeled videos to train our network. These videos are from the OxUvA dataset \cite{2018longtermBenchmark}, where there are 337 videos in total. The OxUvA dataset is a subset of Youtube-BB \cite{youtubeBB} collected on YouTube. By adding these videos during training, our tracker improves the original one by 0.7\% DP and 1.2\% AUC as shown in Table~\ref{table:more training data}.
By leveraging another large-scale LaSOT dataset \cite{LaSOT} where there are 1200 videos collected on the Internet, the tracking performance is further improved. It indicates that unlabeled data advances the unsupervised training. As our framework is fully unsupervised, it has the potential to take advantage of the in-the-wild videos on the Internet to boost the performance.

\setlength{\tabcolsep}{2pt}
\begin{table}
	\scriptsize
	\begin{center}
		\caption{A performance potential of our unsupervised tracker. When using more data (LaSOT) for network training, the performance is further improved. By incorporating empirical features (HOG), our unsupervised tracker achieves superior results. The performance is evaluated on the OTB-2015 dataset using DP and AUC metrics.} \label{table:hand-craft feature}
		\begin{tabular*}{8.0 cm} {@{\extracolsep{\fill}}|l|cccc|}
			\hline
			~ & ECOhc & LUDT+  & LUDT+ &LUDT+   \\
			~ & &only ILSVRC   & more data &more data + HOG \\
			\hline
			\hline
			~DP (\%) &85.4  &84.3 &85.5 &85.8\\
			~AUC (\%)  &64.1 &63.9 &64.9 &65.7\\
			\hline
			~Speed (FPS)  &60 &55 &55 &42\\
			\hline
		\end{tabular*}
	\end{center}
\end{table}

\subsubsection{Empirical Features Embedding}
As shown in Table~\ref{table:hand-craft feature}, we train unsupervised LUDT+ using more unlabeled video sequences (both ILSVRC and LaSOT), which outperforms ECOhc \cite{ECO} leveraging hand-crafted features including HOG \cite{HOG} and ColorName \cite{ColorName}. In addition, we can combine the learned CNN features and empirical features to generate a more discriminative representation. We add the HOG feature to LUDT+ during tracking and evaluate its performance. Table~\ref{table:hand-craft feature} shows that this combination achieves a 65.7\% AUC on OTB-2015. Moreover, embedding the HOG feature helps LUDT+ to outperform most state-of-the-art real-time trackers as shown in Table \ref{table:OTB2015}. Besides feature embedding and adaptive model update, there are still many improvements from \cite{RASNet,BACF,Context-AwareCorrelationFilter,CSR-DCF} available to benefit our tracker. However, adding more additional mechanisms is out of the scope of this work. Following SiamFC and DCFNet, we currently use LUDT/LUDT+ trackers which are \emph{only} trained on the ILSVRC dataset for \emph{fair comparison} in the following evaluations.

\begin{figure}[t]
	\centering
	\includegraphics[width=8.3cm]{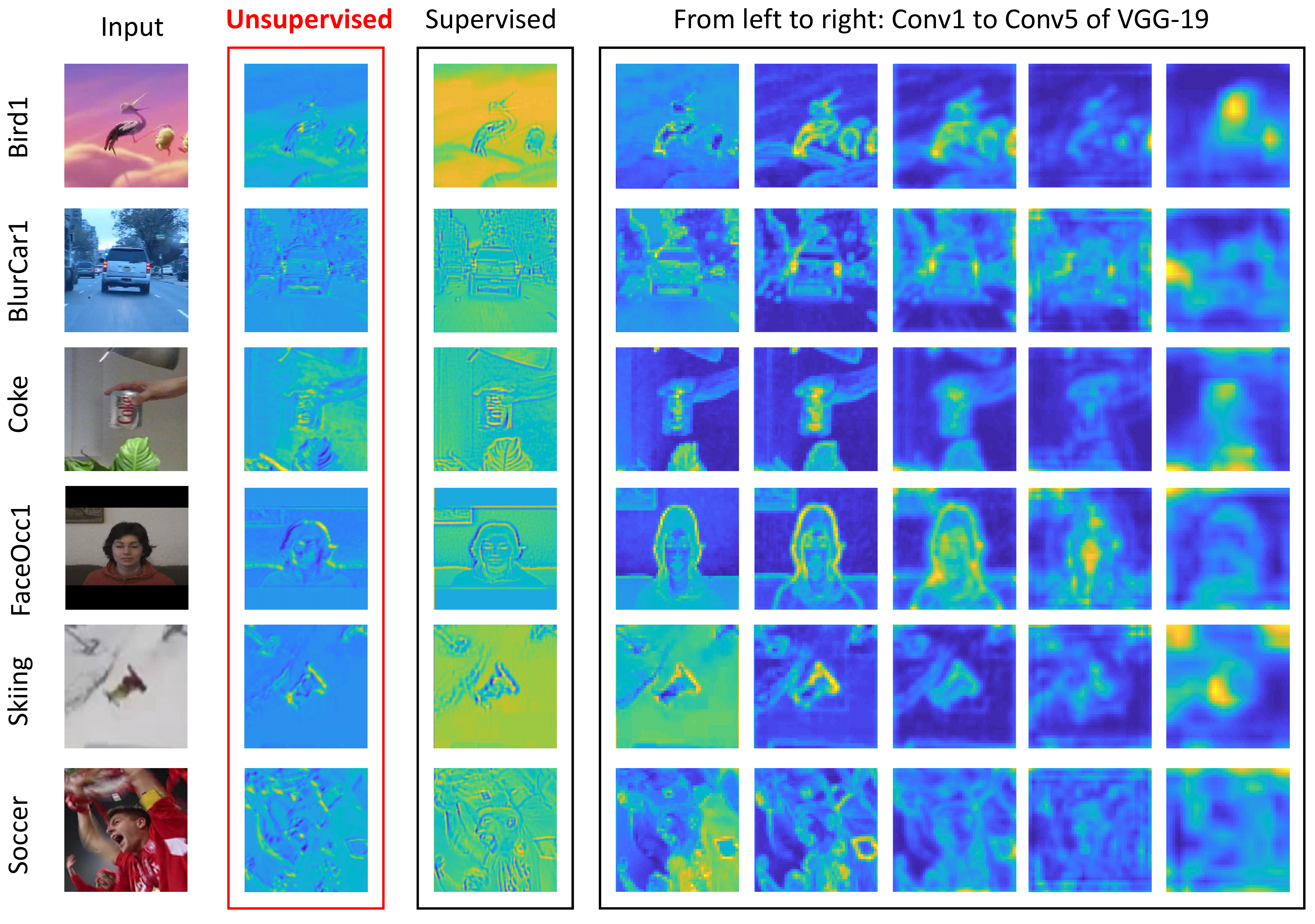}
	\caption{Visualization of feature representations. First column: input image patches. Second and third columns: feature representations of our unsupervised LUDT and fully-supervised LUDT. The rest columns: feature maps from VGG-19 (from left to right: Conv1-2, Conv2-2, Conv3-4, Conv4-4, and Conv5-4). The feature map is visualized by averaging all the channels. Best viewed in color and zoom in.} \label{fig:feature_visualization} 
\end{figure}

\setlength{\tabcolsep}{2pt}
\begin{table*}
	\scriptsize
	\begin{center}
		\caption{Evaluations with fully-supervised baseline (left) and state-of-the-art trackers (right) on the popular OTB-2015 benchmark \cite{OTB-2015}. The evaluation metrics are DP and AUC scores. Our unsupervised LUDT tracker performs favorably against popular baseline methods (left), while our LUDT+ tracker achieves comparable results with the recent state-of-the-art supervised trackers (right).} \label{table:OTB2015}
		\vspace{+0.0in}
		\begin{tabular*}{17.2 cm} {@{\extracolsep{\fill}}|l|cccc|ccccccccccc|}
			\hline
			~Trackers & SiamFC &DCFNet  & CFNet  & {LUDT} & EAST &HP &SA-Siam & SiamRPN &RASNet &SACF &Siam-tri &RT-MDNet &MemTrack & StructSiam & {LUDT+} \\
			\hline
			\hline
			~DP (\%) &77.1  &- &74.8 &76.9 &- &79.6  &86.5  &85.1 &- &83.9 &78.1 &88.5 &82.0 &85.1  &84.3 \\
			~AUC (\%) &58.2  &58.0 &56.8 &60.2 &62.9 &60.1  &65.7  &63.7 &64.2 &63.3 &59.2 &65.0 &62.6 &62.1  &63.9 \\
			\hline
			~FPS  &86  &70 &65 &70 &25 &159 &69  &160 &83 &23 &86 &50 &50 &45  &55\\
			\hline
		\end{tabular*}
	\end{center}
	\vspace{-0.0in}
\end{table*}

\subsection{Visualization of Unsupervised Representation}\label{featuremap}
After learning the unsupervised Siamese tracking network, we visualize the network response to see how it differs from the same network trained using supervised learning.
Fig.~\ref{fig:feature_visualization} shows the visualization performance. The first column shows the input frames. The network responses from unsupervised learning and supervised learning are shown in the second and third columns, respectively. The remaining columns show the feature responses from the off-the-shelf deep model VGG-19 \cite{VGG}. We observe that the responses from the unsupervised learning and supervised learning are similar with minor differences. Specifically, the boundary responses from supervised learning are higher than those of unsupervised learning. This is because of the strong supervisions brought by the ground-truth labels. The network has learned to differentiate the target and background distractors according to labels, which increases the network attention around the object boundaries. In comparison, our unsupervised learning does not employ this process for attention enhancement, while still focusing on the center region of the object responses. From the viewpoint of Siamese network, both unsupervised and supervised feature representations focus on the target appearances, which facilitate the template matching through the correlation operation. Compared with the empirical features (e.g., HOG), we will show in the following that our unsupervised feature representations achieve higher accuracy compared with hand-crafted features.

Our unsupervised representation is compared with the off-the-shelf deep model VGG-19. Note that the VGG model is trained on an image classification task with supervised learning. We show the feature maps from different layers (i.e., Conv1-2, Conv2-2, Conv3-4, Conv4-4, and Conv5-4) of the VGG-19. From Fig.~\ref{fig:feature_visualization}, we observe that our unsupervised feature representations share similarities with the low-level features (i.e., the first two layers) of VGG, which typically represents spatial details. It has been well studied in HCF \cite{HCF} and C-COT \cite{C-COT} that merely using the first or second layer of the VGG model for DCF tracking contains limitations. However, our unsupervised representation better suits the tracking scenario since we jointly combine the feature representation learning with the DCF formulation in an end-to-end fashion. In the deeper layers of VGG-19 such as Conv4-4 and Conv5-4, the feature representation gradually loses spatial details but increases semantics, which can be combined with the low-level features to further boost the tracking performance \cite{HCF,C-COT}. The semantic representation capability is obtained by distinguishing different object categories (i.e., image classification), while our unsupervised learning process lacks such image labels. In the future, we will investigate how to learn rich multiple-level representations for visual tracking under an unsupervised manner.

\subsection{Comparison with State-of-the-art Methods}\label{state-of-the-art}

{\flushleft \bf OTB-2013 Dataset.} The OTB-2013 dataset \cite{OTB-2013} contains 50 challenging videos.
On the OTB-2013 dataset, we evaluate our LUDT and LUDT+ trackers with state-of-the-art real-time trackers including ACT \cite{ACT}, ACFN \cite{ACFN}, CFNet \cite{CFNet}, SiamFC \cite{SiamFc}, SCT \cite{SCT}, CSR-DCF \cite{CSR-DCF}, DSST \cite{DSST}, and KCF \cite{KCF} using precision and success plots.

\begin{figure}
	\centering
	\includegraphics[width=4.0cm]{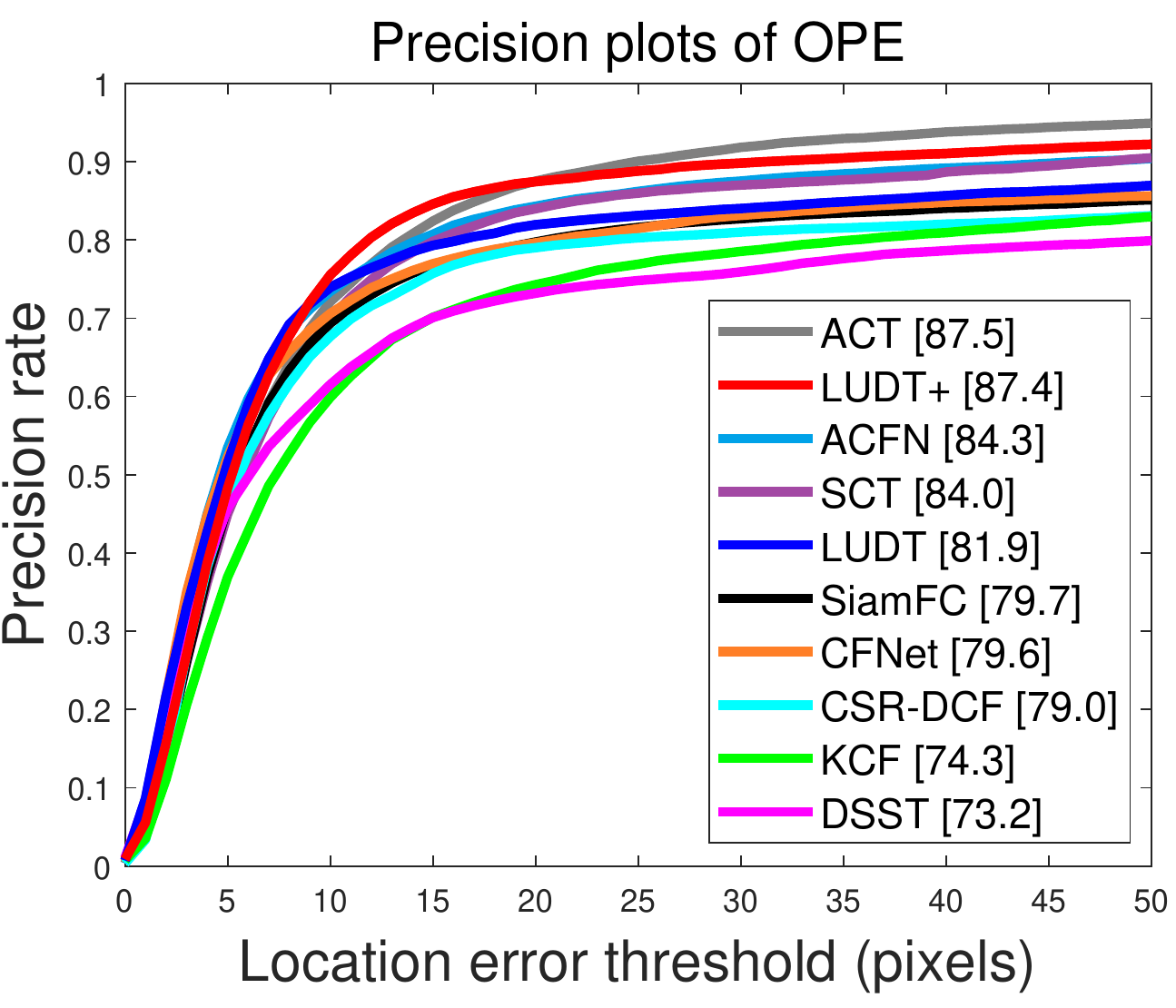}
	\includegraphics[width=4.0cm]{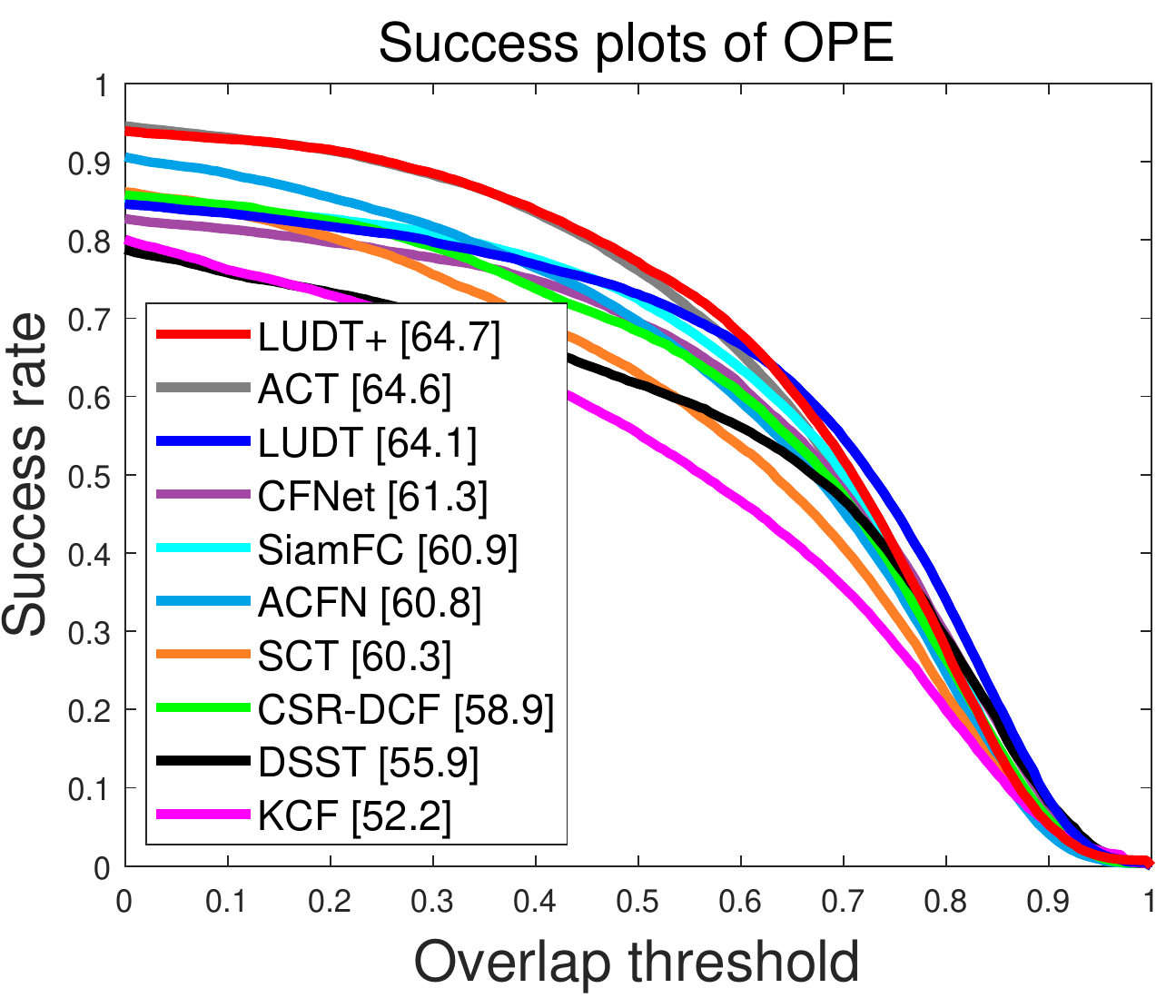}
	\caption{Precision and success plots on the OTB-2013 dataset \cite{OTB-2013} for recent real-time trackers. The legend in each tracker shows the precision at 20 pixels of precision plot and AUC of success plot.} \label{fig:OTB-2013} 
\end{figure}

As illustrated in Fig.~\ref{fig:OTB-2013}, our unsupervised LUDT tracker outperforms CFNet and SiamFC in both distance precision and AUC score. It is worth mentioning that both LUDT and CFNet have similar network capabilities (network depth), leverage the same training data, and are not equipped with additional online improvements. Even though our approach is free of ground-truth supervision, it still achieves very competitive tracking accuracy. Our improved version, LUDT+, performs favorably against recent state-of-the-art real-time trackers such as ACT and ACFN. 
Besides, our LUDT and LUDT+ trackers also exhibit a real-time speed of about 70 FPS and 55 FPS, respectively.

\begin{figure}
	\centering
	\includegraphics[width=4.0cm]{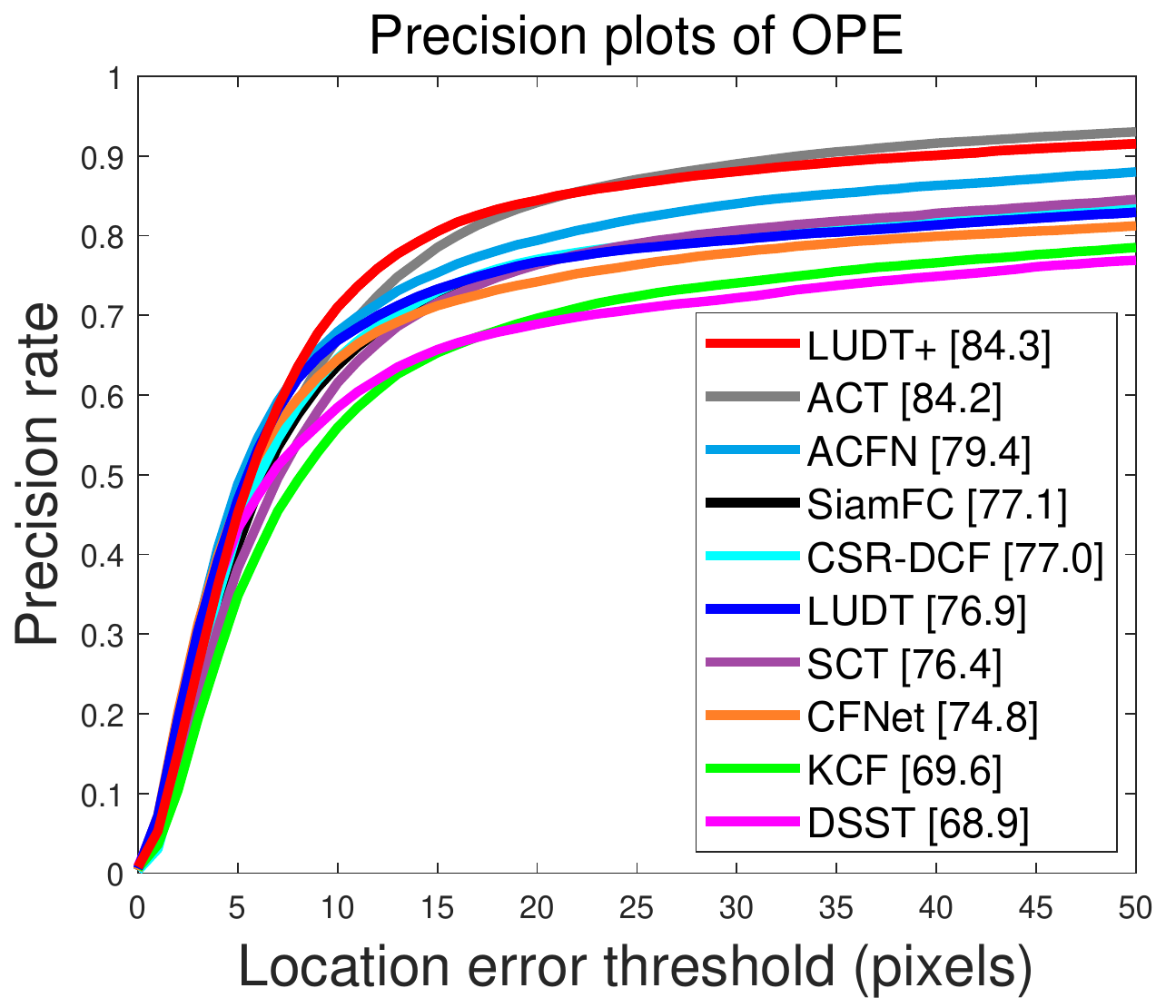}
	\includegraphics[width=4.0cm]{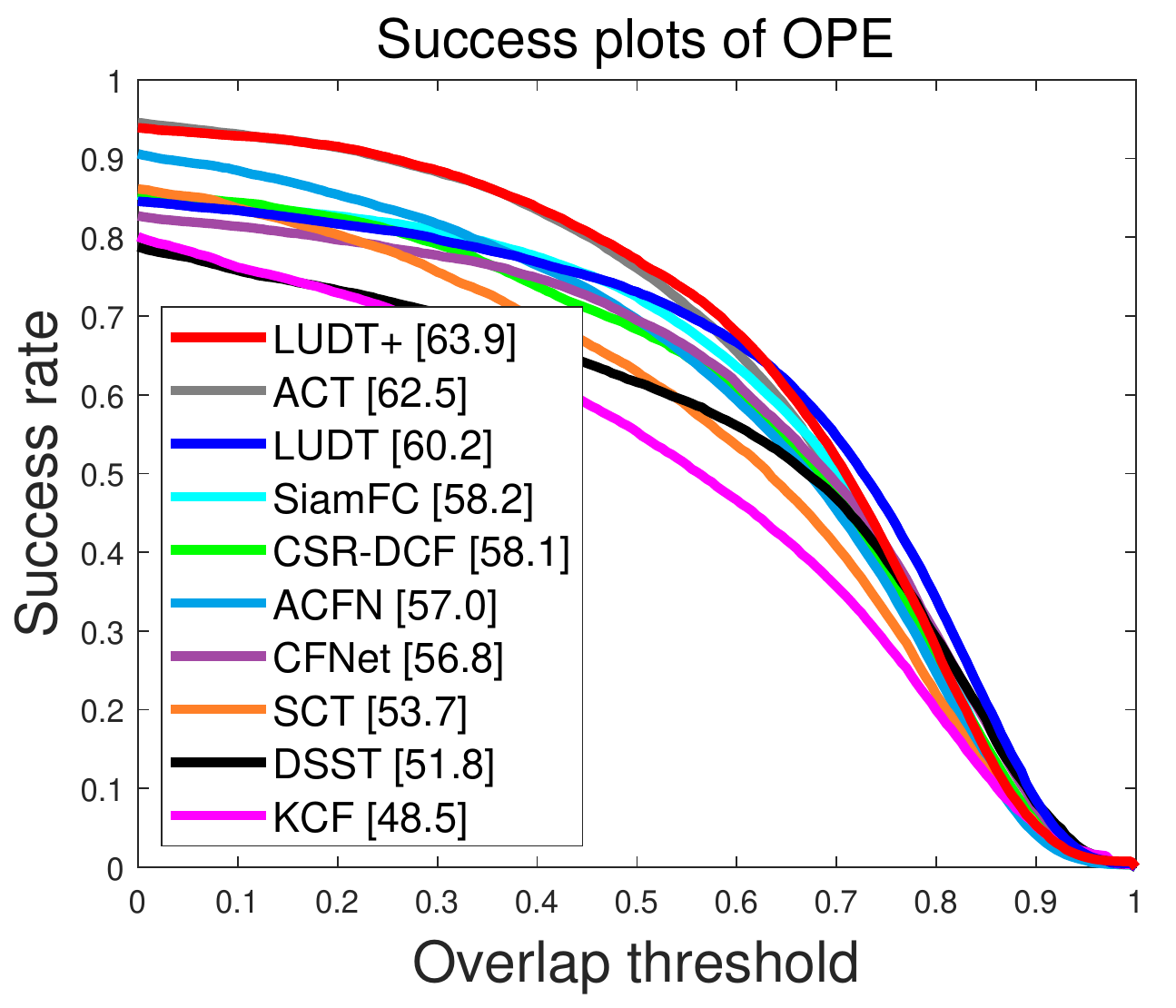}
	\caption{Precision and success plots on the OTB-2015 dataset \cite{OTB-2015} for recent real-time trackers. The legend in each tracker shows the precision at 20 pixels of precision plot and AUC of success plot.} \label{fig:OTB-2015} 
\end{figure}

{\flushleft \bf OTB-2015 Dataset.} The OTB-2015 dataset \cite{OTB-2015} contains 100 challenging videos.
On the OTB-2015 dataset \cite{OTB-2015}, we evaluate LUDT and LUDT+ trackers with state-of-the-art real-time algorithms as that in OTB-2013. In Table~\ref{table:OTB2015}, we further compare our methods with more state-of-the-art real-time trackers such as StructSiam \cite{StructSiam}, MemTrack \cite{MemTrack}, RT-MDNet \cite{RTMDNet}, Siam-tri \cite{Siamtriplet}, SACF \cite{SACF}, RASNet \cite{RASNet}, SiamRPN \cite{SiamRPN}, SA-Siam \cite{SASiam}, HP \cite{HP}, and EAST \cite{EAST}.

From Fig.~\ref{fig:OTB-2015} and Table~\ref{table:OTB2015}, we observe that our unsupervised LUDT tracker is comparable with supervised baseline methods (e.g., SiamFC and CFNet). On the OTB-2015 dataset, SiamFC achieves 77.1\% DP and 58.2\% AUC, while LUDT exhibits 76.9\% DP and 60.2\% AUC. Compared with CFNet, LUDT outperforms by 2.1\% DP and 3.4\% AUC. The DSST algorithm is a traditional DCF based tracker with an accurate target scale estimation. LUDT significantly outperforms it by 8.0\% DP and 8.4\% AUC, which illustrates that our unsupervised feature representation is more robust than empirical features (e.g., HOG).
With a better DCF formulation and more advanced online update strategies \cite{ECO}, our LUDT+ tracker achieves comparable performance with the recent ACFN and ACT trackers.
In Fig.~\ref{fig:OTB-2015} and Table~\ref{table:OTB2015}, we do not compare with some remarkable non-realtime trackers. For example, MDNet \cite{MDNet} and ECO \cite{ECO} can yield 67.8\% and 69.4\% AUC on the OTB-2015, but they are far from achieving a real-time speed.

Table~\ref{table:OTB2015} compares more recent supervised trackers. These latest approaches are mainly based on the Siamese network, which improve the baseline SiamFC method using various sophisticated techniques. Most trackers in Table~\ref{table:OTB2015} are trained using ILSVRC including LUDT+. However, it is worth mentioning that some algorithms (e.g., SA-Siam and RT-MDNet) adopt pre-trained CNN models (e.g., AlexNet \cite{Alexnet} and VGG-M \cite{VGGM}) for network initialization. SiamRPN additionally uses more labeled training videos from the Youtube-BB dataset \cite{youtubeBB}. Compared with them, LUDT+ does not require data labels or off-the-shelf deep models, while still achieving comparable performance and efficiency.

\begin{figure}
	\centering
	\includegraphics[width=4.0cm]{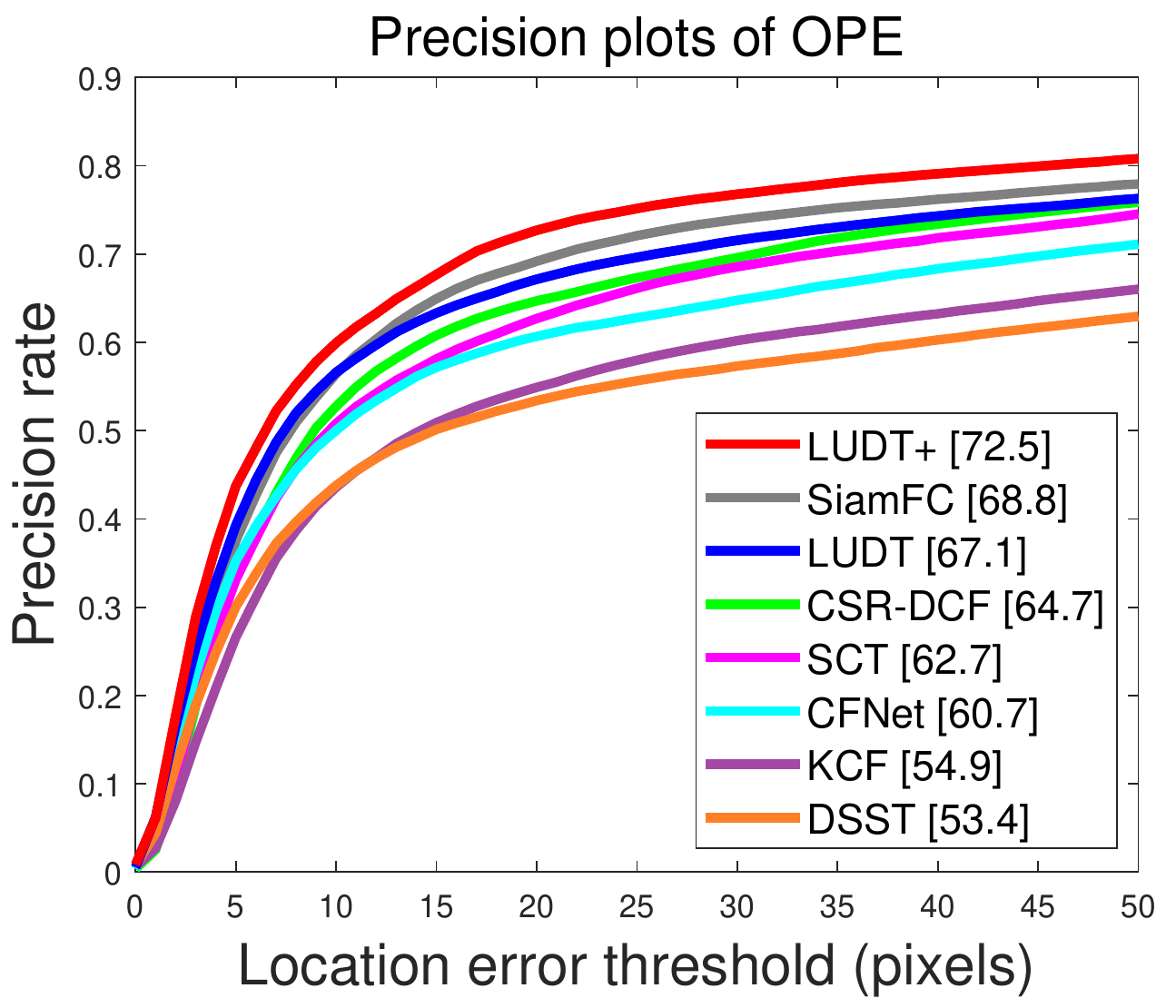}
	\includegraphics[width=4.0cm]{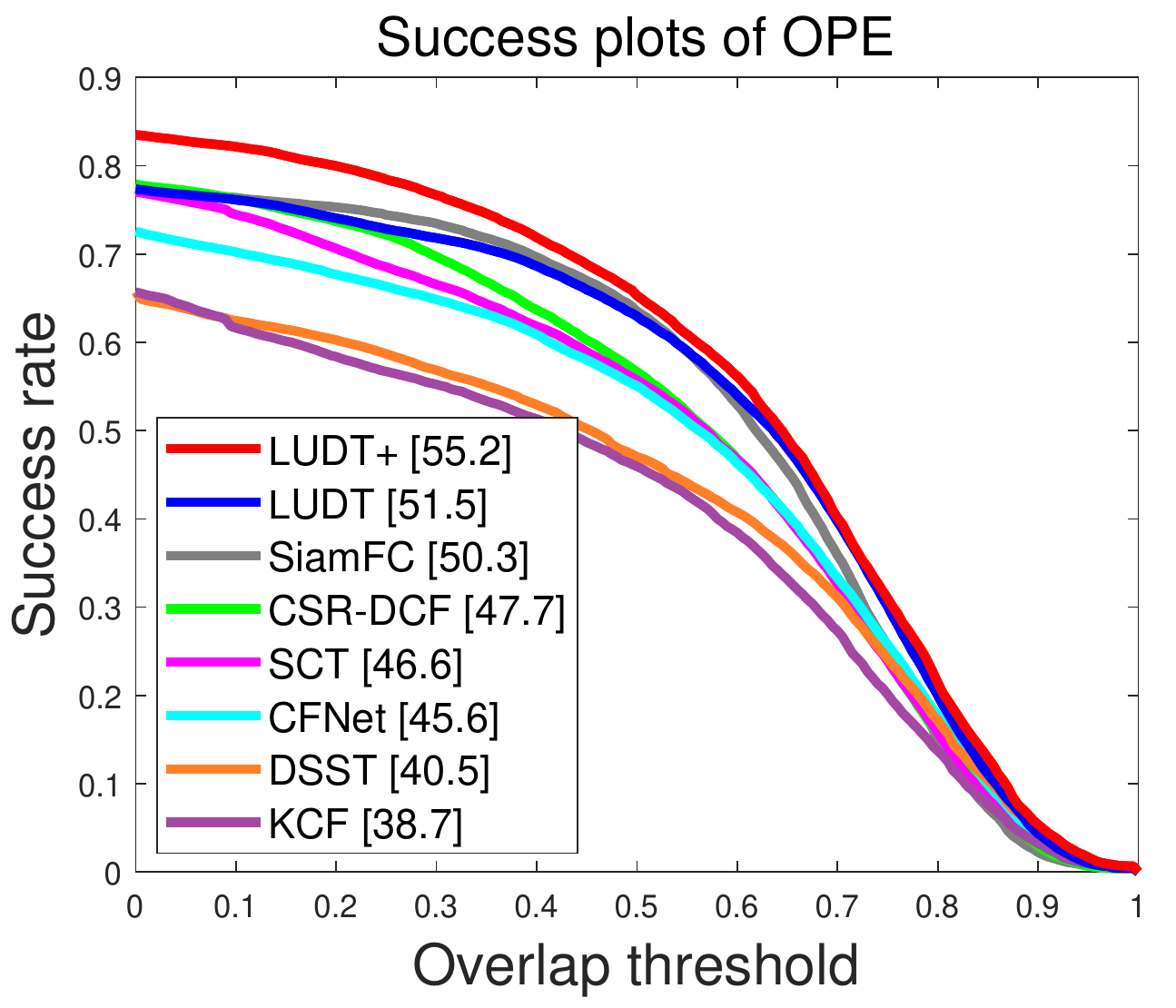}
	\caption{Precision and success plots on the Temple-Color dataset \cite{TempleColor128} for recent real-time trackers. The legend in each tracker shows the precision at 20 pixels of precision plot and AUC of success plot.} \label{fig:Temple-Color} 
\end{figure}

\begin{figure}
	\centering
	\includegraphics[width=4.0cm]{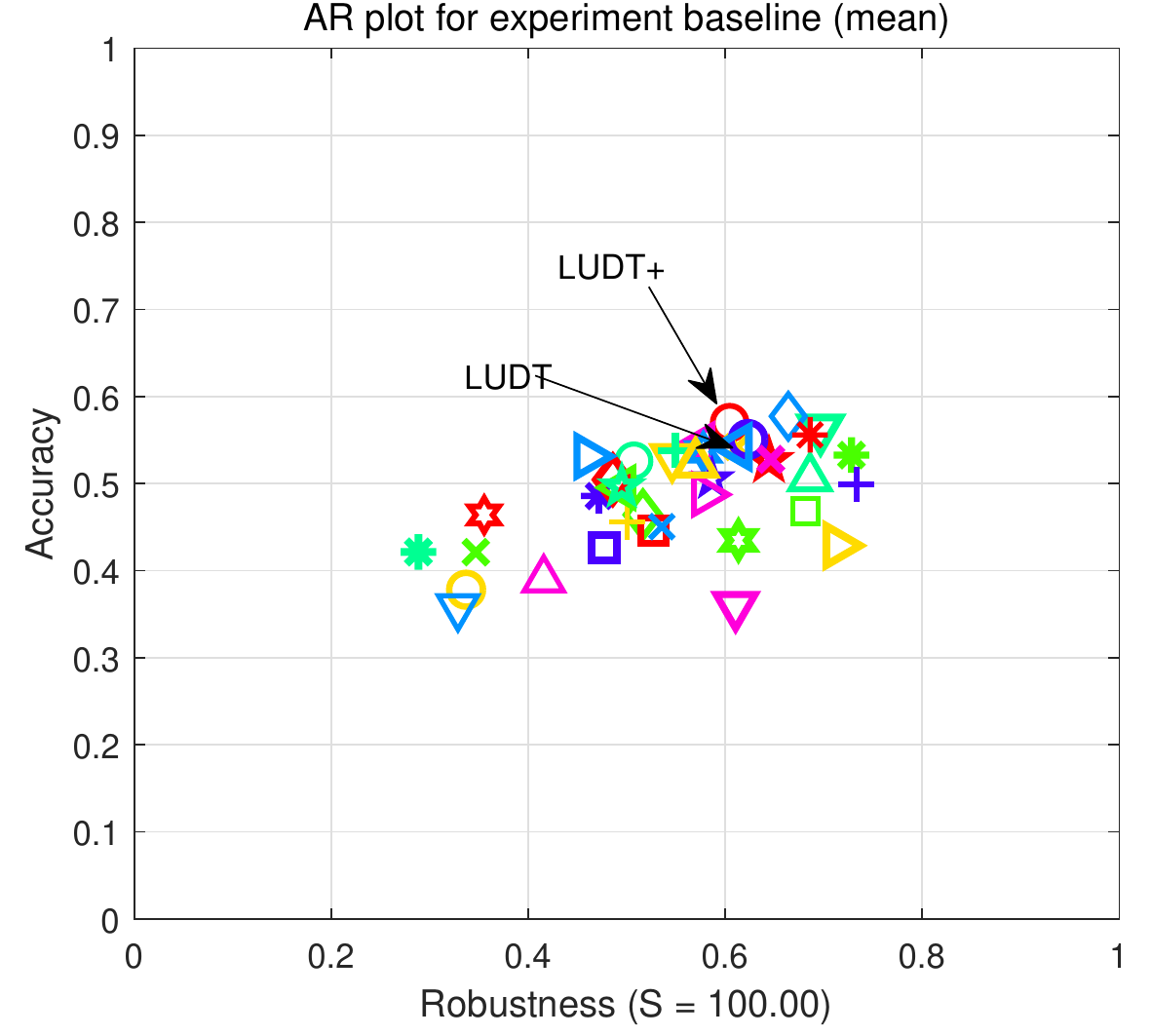}
	\includegraphics[width=4.0cm]{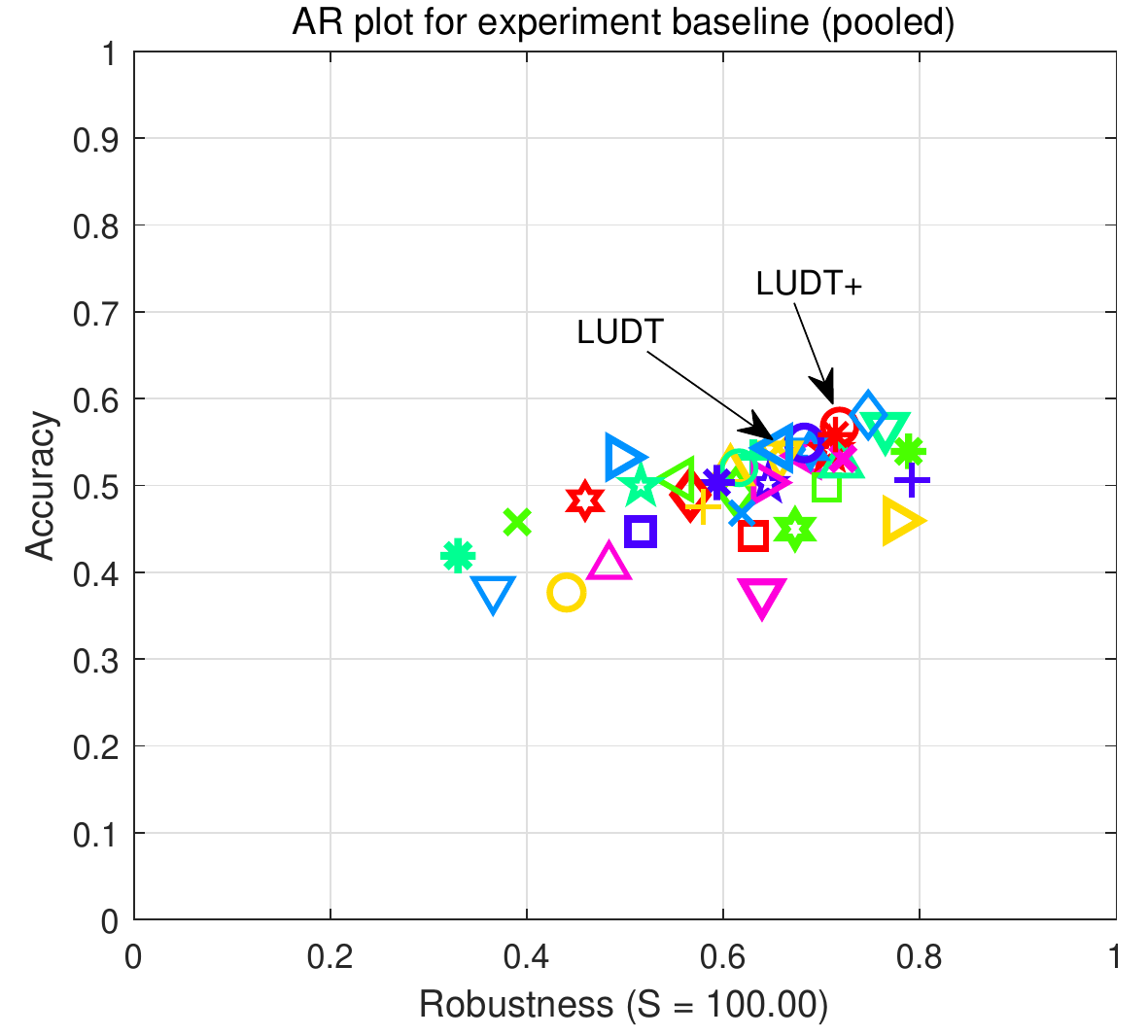}
	\includegraphics[width=8.2cm]{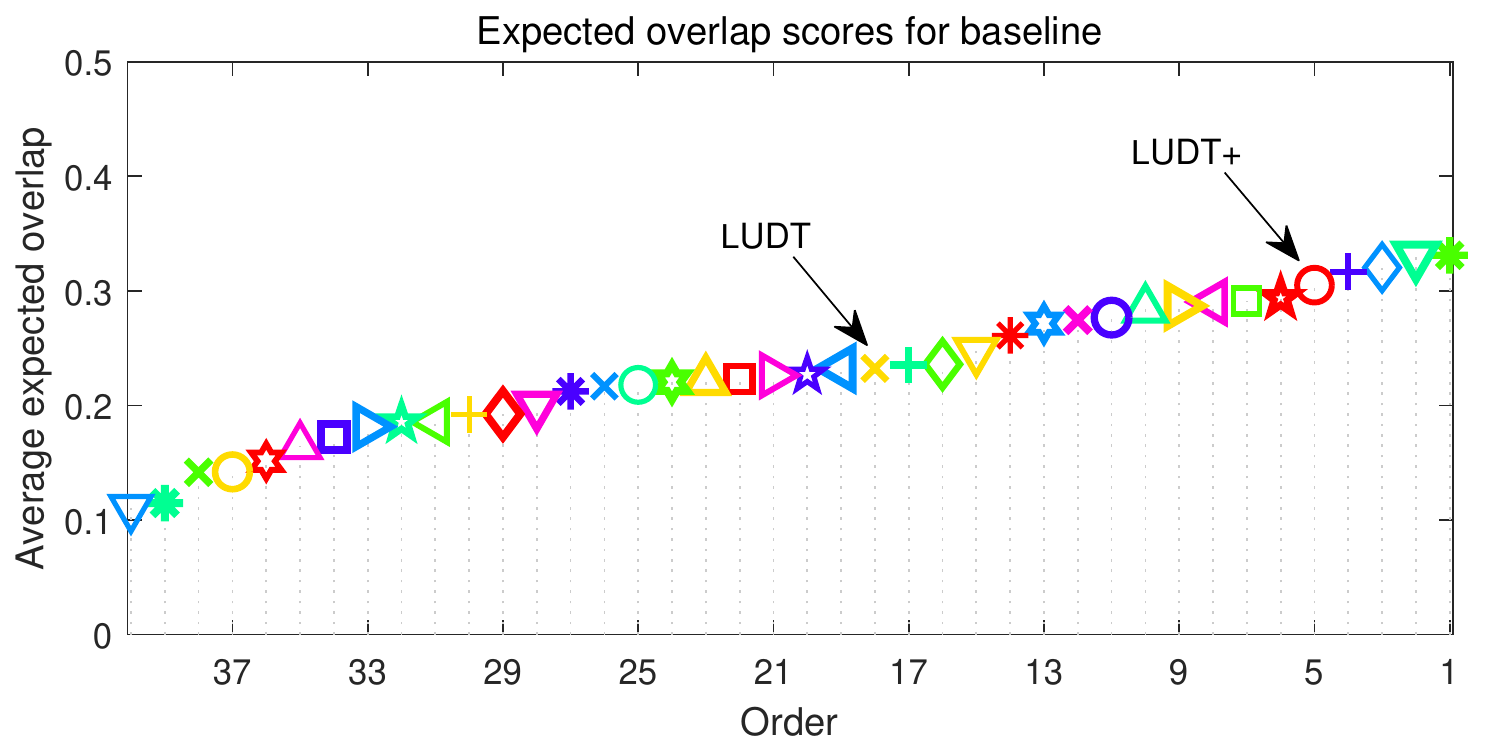}
	\includegraphics[width=8.2cm]{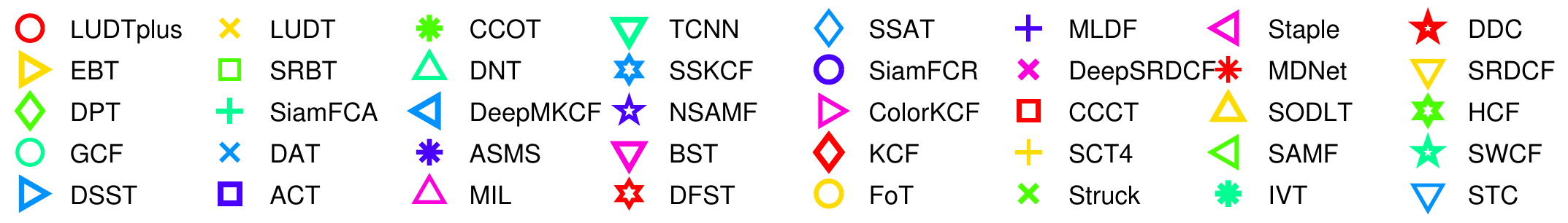}
	\caption{Top: Accuracy-Robustness (AR) ranking plots generated by sequence mean (left) and sequence pooling (right) on the VOT2016 dataset \cite{VOT2016}. Trackers closer to the upper right corner perform better. Bottom: Expected Average Overlap (EAO) graph with trackers ranked from right to left evaluated on VOT2016.}
	\label{fig:VOT2016} 
\end{figure}

\setlength{\tabcolsep}{2pt}
\begin{table}
	\footnotesize
	\begin{center}
		\caption{Comparison with state-of-the-art and baseline trackers on the VOT2016 benchmark \cite{VOT2016}. The evaluation metrics include Accuracy, Failures (over 60 sequences), and Expected Average Overlap (EAO). The up arrows indicate that higher values are better for the corresponding metric and vice versa. } \label{table:VOT2016}	
		\begin{tabular*}{7.9 cm} {@{\extracolsep{\fill}}|l|ccc|c|}
			\hline
			~~Trackers & Accuracy ($ \uparrow $) & Failures ($ \downarrow $) & EAO ($ \uparrow $)~ & FPS ($ \uparrow $)~ \\
			\hline
			\hline
			~~ECO  &0.54 &- &0.374 &6\\
			~~VITAL  &- &- &0.323 &1 \\
			~~DSLT & - &- & 0.332 & 6\\
			~~RTINet &0.57 &- &0.298 &9 \\
			~~C-COT &0.52 &51 &0.331 & 0.3\\
			~~pyMDNet &- & - & 0.304 & 2\\
			~~HCF &0.45 &85 &0.220 &12\\
			~~ACT & - &- & 0.275 &30 \\
			~~SA-Siam &0.53 &- &0.291 & 50\\
			~~SiamRPN  &0.56 &- &0.344 &160\\
			~~SACF & - &- & 0.275 &23\\
			~~StructSiam &- &- &0.264 &45\\
			~~MemTrack & 0.53 & - & 0.273 &50\\
			~~SiamFC  &0.53  &99 &0.235 &86\\	
			~~SCT4 &0.48 &117 &0.188 &40\\
			~~DSST &0.53 &151 &0.181&25\\	
			~~KCF &0.49 &122 &0.192 &170\\
			\hline
			\hline
			~~LUDT (Ours) &0.54 &100 &0.231 &70\\
			~~LUDT+ (Ours) &0.54 &62 &0.309 &55\\
			\hline
		\end{tabular*}
	\end{center}
\end{table}

{\flushleft \bf Temple-Color Dataset.} Temple-Color \cite{TempleColor128} is a more challenging benchmark with 128 color videos.
In this dataset, we compare our trackers with some baselines and state-of-the-art trackers as on the OTB-2015 benchmark.
Compared with the DCF trackers with empirical features (e.g., HOG feature), our tracker with unsupervised deep features exhibits a significant performance improvement as shown in Fig.~\ref{fig:Temple-Color}.
Specifically, SiamFC which is learned with full supervision achieves an AUC score of 50.3\%, while LUDT exhibits a 51.3\% AUC score.
Compared with another representative supervised method CFNet, LUDT exceeds its performance by 6.4\% DP and 4.7\% AUC.
Furthermore, our LUDT+ tracker performs favorably against existing state-of-the-art trackers.

{\flushleft \bf VOT2016 Dataset.} We report the evaluation results on the VOT2016 benchmark \cite{VOT2016}, which contains 60 videos selected from more than 300 videos.
Different from the OTB dataset, the VOT toolkit will reinitialize when the tracker fails.
The expected average overlap (EAO) is the final metric for tracker ranking \cite{VOTpami}.
In Fig.~\ref{fig:VOT2016}, we show the accuracy-robustness (AR) plot and EAO ranking plot on VOT2016 with some participant trackers. The VOT2016 champion C-COT uses the pre-trained VGG-M model for feature extraction while not achieving real-time performance.
The proposed LUDT+ method performs slightly worse than C-COT but runs much faster.
It is worth mentioning that our real-time LUDT+ tracker even performs favorably against remarkable non-realtime deep trackers such as MDNet. Our LUDT tracker, without bells and whistles, surpasses classic DCF trackers such as DSST and KCF by a considerable margin and is comparable with some DCF methods with an off-the-shelf deep model (e.g., DeepMKCF and HCF).

In Table~\ref{table:VOT2016}, we include more state-of-the-art trackers including VITAL \cite{VITAL}, DSLT \cite{DSLT}, RTINet \cite{RTINet}, ACT \cite{ACT}, SA-Siam \cite{SASiam}, SiamRPN \cite{SiamRPN}, SACF \cite{SACF}, StructSiam \cite{StructSiam}, and MemTrack \cite{MemTrack} on the VOT2016 benchmark. 
Compared with the baseline SiamFC, our LUDT tracker yields favorable results. Compared with fully-supervised trackers, LUDT+ overall exhibits competitive performance as well as efficiency.

\setlength{\tabcolsep}{2pt}
\begin{table}
	\footnotesize
	\begin{center}
		\caption{Comparison with state-of-the-art and baseline trackers on the VOT2017/2018 benchmark \cite{VOT2016}. The evaluation metrics include Accuracy, Failures (over 60 sequences), and Expected Average Overlap (EAO). The up arrows indicate that higher values are better for the corresponding metric and vice versa. } \label{table:VOT2017}	
		\begin{tabular*}{7.9 cm} {@{\extracolsep{\fill}}|l|ccc|c|}
			\hline
			~~Trackers & Accuracy ($ \uparrow $) & Failures ($ \downarrow $) & EAO ($ \uparrow $)~ & FPS ($ \uparrow $)~ \\
			\hline
			\hline
			~~ECO &0.48 &59 &0.280 &6\\
			~~C-COT  &0.49 &68 &0.267 & 0.3\\
			~~SA-Siam &0.50 &- &0.236 & 50\\
			~~SiamRPN &- &- &0.243 &160\\
			~~SiamFC &0.50  &125 &0.188 &86\\	
			~~Staple &0.53  &147 &0.169 &70\\	
			~~TRACA &0.42  &183 &0.137 &100\\	
			~~SRDCF &0.49  &208 &0.119 &5\\	
			~~DSST &0.40 &310 &0.079 &25\\	
			~~KCF &0.45 &165 &0.135 &170\\
			\hline
			\hline
			~~LUDT (Ours) &0.46 &149 &0.154 &70\\
			~~LUDT+ (Ours) &0.49 &88 &0.230 &55\\
			\hline
		\end{tabular*}
	\end{center}
\end{table}

\begin{figure}
	\centering
	\includegraphics[width=7.9cm]{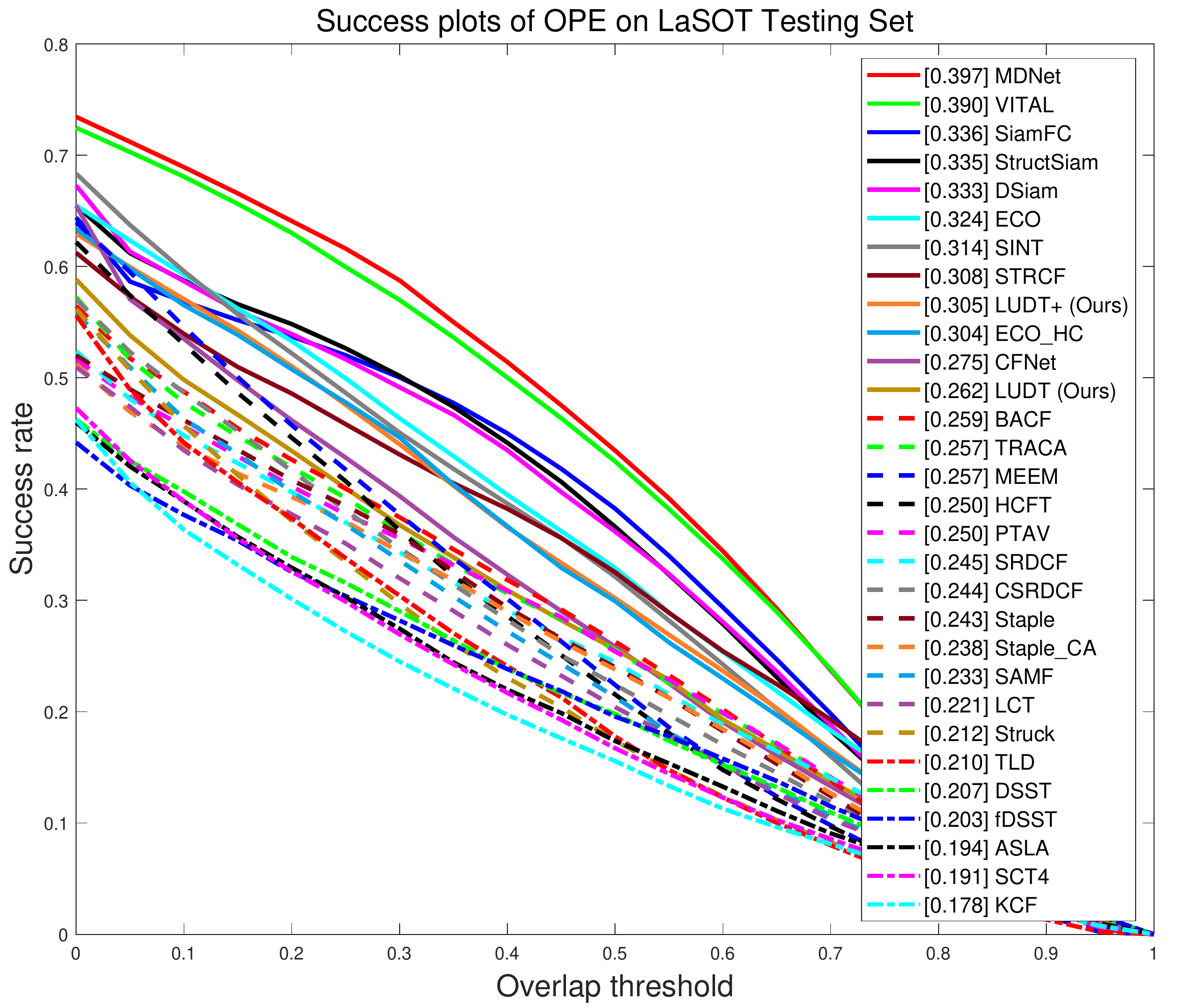}
	\caption{Success plots on the LaSOT testing set \cite{LaSOT}. The legend in each tracker shows the AUC of the success plot. Best viewed in color and zoom in.} \label{fig:lasot} 
\end{figure}

\setlength{\tabcolsep}{2pt}
\begin{table}
	\footnotesize
	\begin{center}
		\caption{Comparison with state-of-the-art and baseline trackers on the TrackingNet benchmark \cite{2018trackingnet}. The evaluation metrics include Precision, Normalized Precision, and Success (AUC score).} \label{table:trackingnet}	
		\begin{tabular*}{7.6 cm} {@{\extracolsep{\fill}}|l|ccc|}
			\hline
			~~Trackers & Precision & Norm.Prec & Success \\
			\hline
			\hline
			~~MDNet &0.565 &0.705 &0.606 \\
			~~CFNet &0.533 &0.654 &0.578 \\
			~~SiamFC  &0.533 &0.663 &0.571 \\
			~~ECO &0.492 &0.618 &0.554 \\
			~~ECOhc &0.476 &0.608 &0.541 \\
			~~CSRDCF  &0.480 &0.622 &0.534 \\
			~~Staple\_CA &0.468 &0.605 &0.529 \\
			~~Staple &0.470 &0.603 &0.528 \\
			~~BACF &0.461 &0.580 &0.523 \\
			~~SRDCF &0.455 &0.573 &0.521 \\
			~~SAMF &0.477 &0.598 &0.504 \\
			~~ASLA &0.406 &0.536 &0.478 \\
			~~SAMF\_AT &0.447 &0.560 &0.472 \\
			~~DLSSVM &0.418 &0.562 &0.470 \\
			~~DSST &0.460 &0.588 & 0.464 \\
			~~MEEM &0.386 &0.545 &0.460 \\
			~~Struck &0.402 &0.539 &0.456 \\
			~~DCF &0.419 &0.548 & 0.448 \\
			~~KCF &0.419 &0.546 &0.447 \\
			~~CSK &0.368 &0.503 & 0.429 \\
			~~TLD &0.336 &0.460 &0.417 \\
			~~TLD &0.292 &0.438 &0.400 \\
			~~MOSSE &0.326 &0.442 & 0.388 \\
			\hline
			\hline
			~~LUDT (Ours) &0.469 &0.593 &0.543 \\
			~~LUDT+ (Ours) &0.495 &0.633 &0.563 \\
			\hline
		\end{tabular*}
	\end{center}
\end{table}

{\flushleft \bf VOT2017/2018 Dataset.} The VOT2017 \cite{VOT2017} and VOT2018 \cite{VOT2018} are the same benchmark with more challenging videos compared with those in the VOT2016 dataset.
In Table~\ref{table:VOT2017}, we present the Accuracy, Failures, and EAO of the state-of-the-art trackers on VOT2017/VOT2018.
The proposed LUDT tracker is still superior to the standard DCF trackers using hand-crafted features such as DSST and KCF.
Our LUDT+ yields an EAO score of 0.230, which is comparable with the advanced Siamese trackers such as SA-Siam and SiamRPN that take advantage of additional backbone networks or  training data.

\begin{figure*}
	\centering
	\includegraphics[width=17.2cm]{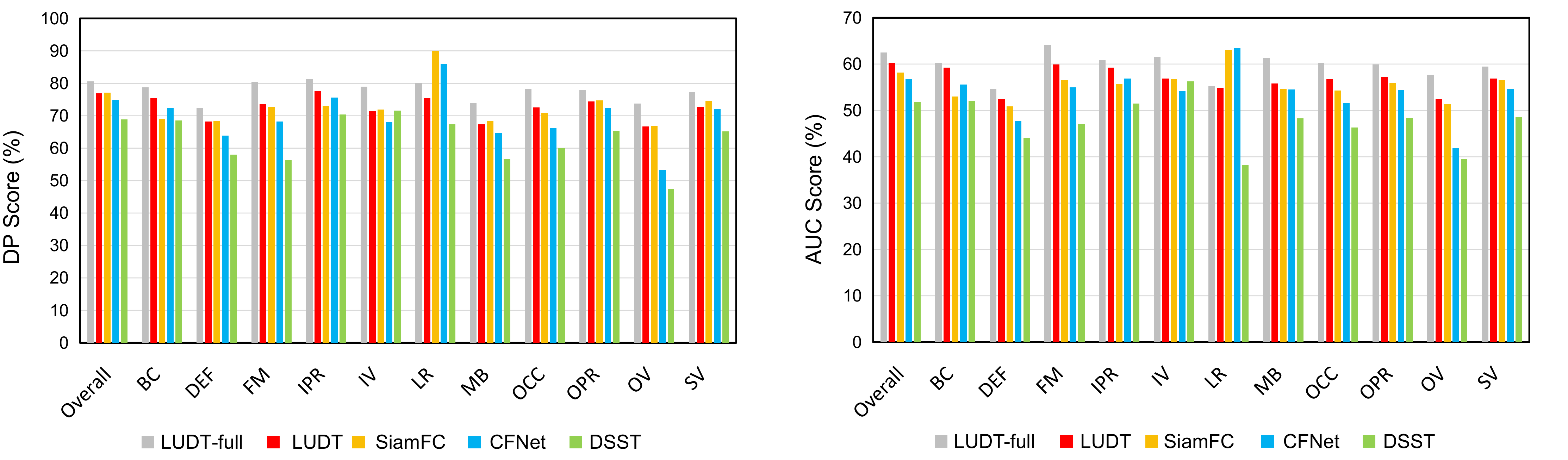}
	\caption{Attribute-based evaluation on the OTB-2015 dataset \cite{OTB-2015}. The 11 attributes are background clutter (BC), deformation (DEF), fast motion (FM), in-plane rotation (IPR), illumination variation (IV), low resolution (LR), motion blur (MB), occlusion (OCC), out-of-plane rotation (OPR), out-of-view (OV), and scale varition (SV), respectively.}
	\label{fig:attribute}
\end{figure*}

\begin{figure*}
	\centering
	\includegraphics[width=17.2cm]{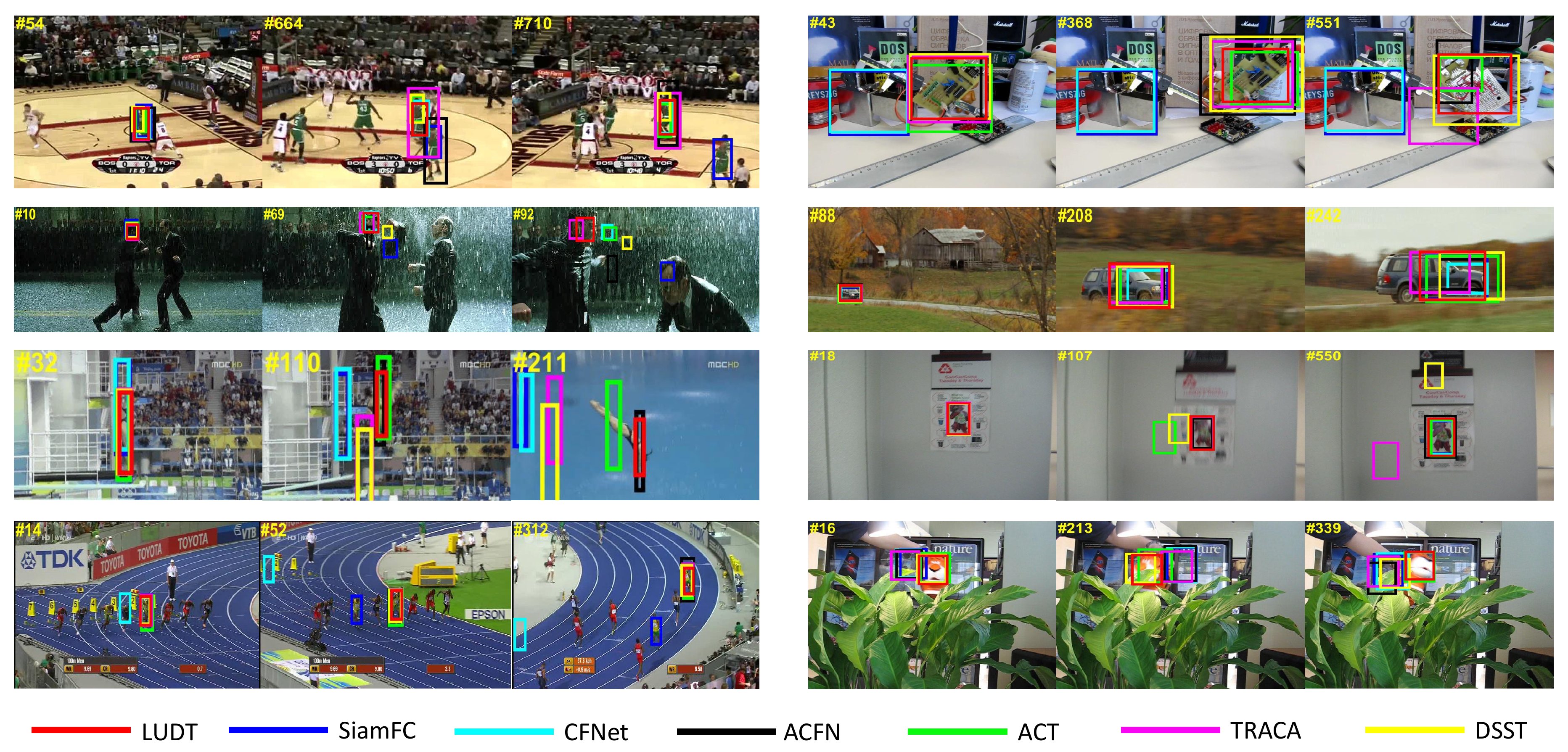}
	\caption{Qualitative evaluation of our proposed LUDT and other trackers including SiamFC \cite{SiamFc}, CFNet \cite{CFNet}, ACFN \cite{ACFN}, ACT \cite{ACT}, TRACA \cite{TRACA}, and DSST \cite{DSST} on 8 challenging videos from OTB-2015. From left to right and top to down are \emph{Basketball}, \emph{Board}, \emph{Matrix}, \emph{CarScale}, \emph{Diving}, \emph{BlurOwl}, \emph{Bolt}, and \emph{Tiger1}, respectively. Best viewed in color.} \label{fig:tracking results} 
\end{figure*}

{\flushleft \bf LaSOT Dataset.} We further evaluate our unsupervised approach on the large-scale LaSOT testing dataset \cite{LaSOT} with 280 videos. The videos in LaSOT are more challenging with an average length of about 2500 frames.
As shown in Fig.~\ref{fig:lasot}, our LUDT tracker still outperforms hand-crafted feature based DCF trackers such as BACF \cite{BACF}, CSR-DCF \cite{CSRDCF_IJCV}, DSST \cite{DSST}, and SCT4 \cite{SCT}.
Furthermore, the proposed LUDT+ approach achieves an AUC score of 30.5\%, which is even comparable with some state-of-the-art deep DCF trackers including ECO (32.4\%) \cite{ECO}, STRCF (30.8\%) \cite{STRCF}, and TRACA (25.7\%) \cite{TRACA} that leverage off-the-shelf deep models as feature extractors.

{\flushleft \bf TrackingNet Dataset.} The recently released large-scale TrackingNet dataset \cite{2018trackingnet} contains more than 30K videos with more than 14 million dense bounding box annotations. The videos are collected on the Internet (YouTube), providing large-scale high-quality data for assessing trackers in the wild. We test our LUDT and LUDT+ on the testing set with 511 videos. Following \cite{2018trackingnet}, we adopt three metrics including Precision, Normalized Precision, and Success (AUC) for performance evaluation. In Table \ref{table:trackingnet}, we exhibit the results of our methods and all the evaluated trackers on this benchmark. On this dataset, our LUDT achieves an AUC score of 54.3\%, which obviously outperforms other hand-crafted feature based DCF trackers such as ECOhc, CSR-DCF, and BACF by 0.2\%, 0.9\%, and 2.0\%, respectively. Note that the above DCF trackers are improved versions with additional regularization terms, while ours merely utilizes a standard DCF formulation.
Our superior performance illustrates the representational power of our unsupervised features.
Besides, it is worth mentioning that our LUDT, on this large-scale benchmark, is even comparable with the state-of-the-art ECO, which leverages both hand-crafted and off-the-shelf deep features. Without labeled data for model training, our improved LUDT+ achieves better performance and slightly outperforms ECO by 0.9\% in terms of AUC.

{\flushleft \bf Attribute Analysis.} The videos on OTB-2015 \cite{OTB-2015} are annotated with 11 different attributes, namely: background clutter (BC), deformation (DEF), out-of-plane rotation (OPR), scale variation (SV), occlusion (OCC), illumination variation (IV), motion blur (MB), in-plane rotation (IPR), out-of-view (OV), fast motion (FM), and low resolution (LR). On the OTB-2015 benchmark, we further analyze the performances over different challenges in Fig.~\ref{fig:attribute}. On the majority of challenging scenes, our LUDT tracker outperforms the popular SiamFC \cite{SiamFc} and CFNet \cite{CFNet} trackers. However, our performance advantage in the DP metric is less obvious than that in the AUC metric. Compared with the fully-supervised LUDT tracker, the main performance gaps are from illumination variation (IV), occlusion (OCC), and fast motion (FM) attributes. Unsupervised learning can be further improved on these attributes.

{\flushleft \bf Qualitative Evaluation.} We evaluate LUDT with supervised trackers (e.g., ACT, ACFN, SiamFC, TRACA, and CFNet) and a baseline DCF tracker (DSST) on eight challenging videos, as shown in Fig.~\ref{fig:tracking results}.
On \emph{Matrix} and \emph{Tiger1} videos, the targets undergo partial occlusion and background clutter, while on \emph{BlurOwl}, the target is extremely blurry due to the drastic camera shaking.
In these videos, DSST based on empirical features fails to cope with the challenging factors while LUDT is able to handle. This illustrates the robustness of our unsupervised feature representation, which achieves favorable performance compared to the empirical features. The SiamFC and CFNet trackers tend to drift when the target and distractors are similar (e.g., \emph{Bolt} and \emph{Basketball} sequences) while LUDT is able to handle these challenging scenes because of the discriminating capability of DCF and its online model update mechanism. 
%
Without online improvements, LUDT is still able to track the target accurately, especially on the challenging \emph{Board} and \emph{Diving} videos.
It is worth mentioning that such a robust tracker is learned from raw videos under an unsupervised manner.

\subsection{Limitations}\label{Limitation}
\begin{figure}
	\centering
	\includegraphics[width=8.3cm]{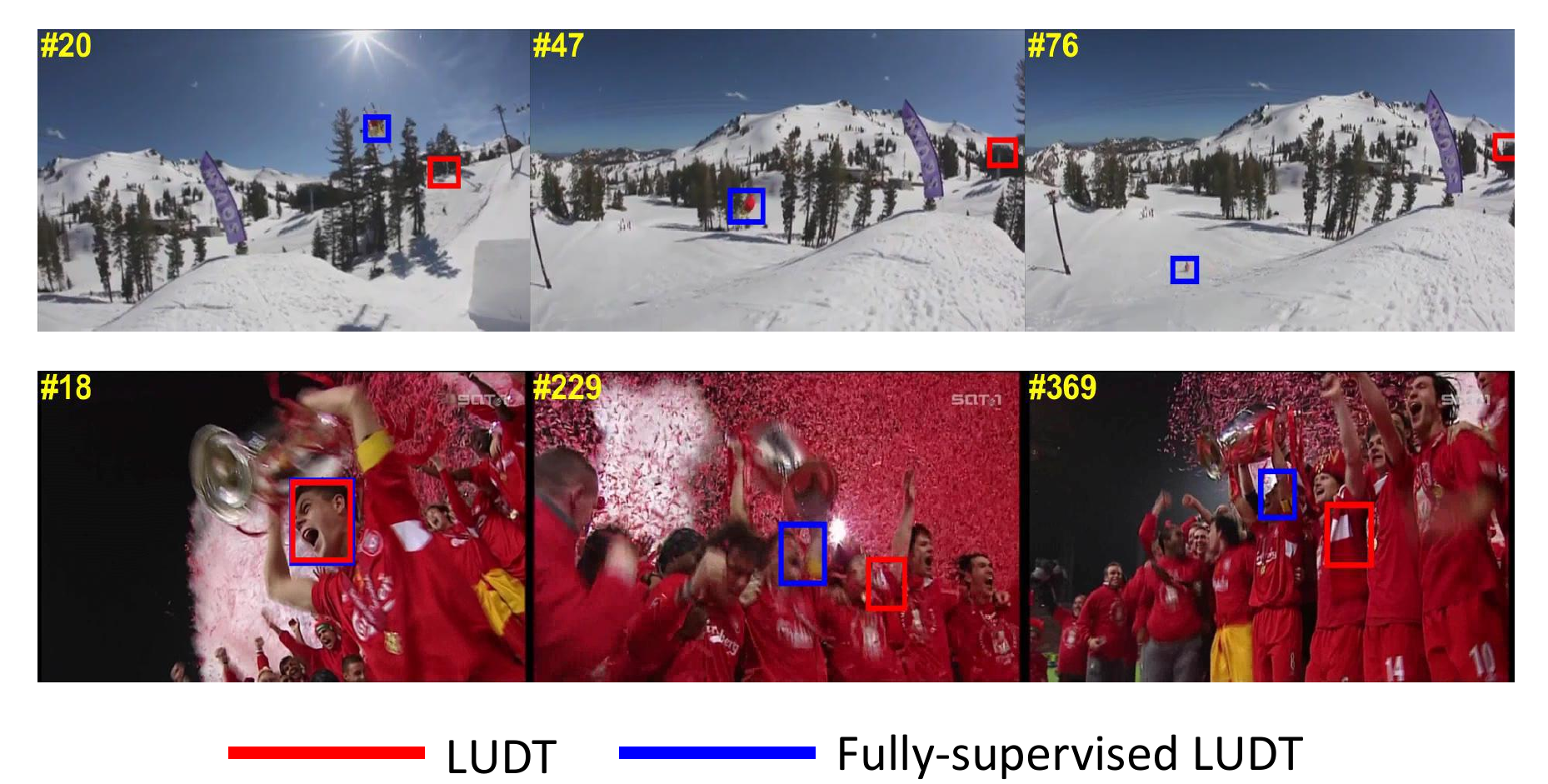}
	\caption{Failure cases of our LUDT tracker. The top and bottom videos are \emph{Skiing} and \emph{Soccer}, respectively. Compared to its fully-supervised version, our unsupervised method is not robust enough when the target undergoes drastic appearance change and occlusion.}
	\label{fig:failure cases}
\end{figure}

Fig.~\ref{fig:failure cases} shows the limitations of our unsupervised learning. First, compared with the fully supervised learning, our tracker trained via unsupervised learning tends to drift when an occlusion or a drastic appearance change occurs (e.g., the targets in \emph{Skiing} and \emph{Soccer} sequences). The semantic representations brought by ground-truth annotations are missing. Second, our unsupervised learning involves both forward and backward trackings. The computational load during the training phase is a potential drawback although the learning process is offline.

\section{Concluding Remarks}\label{conclusion}
In this paper, we present how to train a visual tracker using unlabeled videos in the wild, which is rarely investigated in visual tracking.
By designing an unsupervised Siamese correlation filter network, we verify the feasibility and effectiveness of our forward-backward based unsupervised training pipeline.
To further facilitate the unsupervised training, we extend our framework to consider multiple frames and employ a cost-sensitive loss.
Extensive experiments exhibit that the proposed unsupervised tracker, without bells and whistles, performs as a solid baseline and achieves comparable results with the classic fully-supervised trackers.
Equipped with additional online improvements such as a sophisticated update scheme, our LUDT+ tracker performs favorably against the state-of-the-art tracking algorithms.
Furthermore, we provide a deep analysis of our unsupervised representation by feature visualization and extensive ablation studies.
Our unsupervised framework shows a promising potential in visual tracking, such as utilizing more unlabeled data or weakly labeled data to further improve the tracking accuracy.

\begin{acknowledgements}
This work was supported in part to Dr. Houqiang Li by NSFC under contract No. 61836011, and in part to Dr. Wengang Zhou by NSFC under contract No. 61822208 \& 61632019 and Youth Innovation Promotion Association CAS (No. 2018497). Dr. Chao Ma was supported by NSFC under contract No. 60906119 and Shanghai Pujiang Program.
\end{acknowledgements}

%
%

\bibliographystyle{spmpsci}      
\bibliography{reference}


\end{document}